\documentclass[11pt,letterpaper]{article}

\usepackage[margin=1in]{geometry}

\usepackage{fontspec}
\usepackage{amsmath}
\usepackage{amssymb}

\usepackage{graphicx}
\graphicspath{{./}}

\usepackage{booktabs}
\usepackage{array}
\usepackage{longtable}
\usepackage{multirow}
\usepackage{pdflscape}
\usepackage{placeins}   

\usepackage{xcolor}
\usepackage{microtype}

\usepackage[font=small,labelfont=bf,textfont=it]{caption}

\usepackage[bottom]{footmisc}

\numberwithin{figure}{section}
\numberwithin{table}{section}
\makeatletter
\newenvironment{hangparas}[2]{%
  \par\@afterindenttrue   
  \setlength{\parindent}{-#1}%
  \setlength{\leftskip}{#1}%
  \setlength{\parskip}{0.45\baselineskip}%
}{\par}
\makeatother
\newcommand{\authdash}{\leavevmode\rule[0.45ex]{2em}{0.45pt}}

\usepackage[
  colorlinks=true,
  linkcolor=blue!60!black,
  citecolor=blue!60!black,
  urlcolor=blue!60!black,
  bookmarksnumbered=true
]{hyperref}
\usepackage{xurl}   
\hypersetup{
  pdftitle={MiqraBERT: Regression-Based Sentence-BERT Finetuning for Biblical Hebrew Parallel Detection},
  pdfauthor={David M. Smiley},
  pdfsubject={Biblical Hebrew NLP; semantic textual similarity; parallel detection}
}

\usepackage{polyglossia}
\setmainlanguage{english}
\setotherlanguage{hebrew}
\newfontfamily\hebrewfont{SBL_Hbrw.ttf}[Path=fonts/, Script=Hebrew]

\title{\textbf{MiqraBERT}: Regression-Based Sentence-BERT Finetuning\\
       for Biblical Hebrew Parallel Detection}
\author{David M. Smiley\\[2pt]
  \normalsize University of Notre Dame\\
  \normalsize \texttt{dsmiley@nd.edu}}
\date{}

\begin{document}
\maketitle

\begin{abstract}
\noindent
Textual reuse pervades the Hebrew Bible, yet the computational methods used to
detect it still rest largely on lexical overlap, and they falter once a parallel
involves paraphrase, lexical substitution, or syntactic reworking. This paper
introduces MiqraBERT, a Sentence-BERT model finetuned from AlephBERT (a Modern
Hebrew encoder) for verse-level semantic similarity in Biblical Hebrew. The
training set comprises 1{,}650 labeled verse and half-verse pairs: 825 true
parallels drawn from the Chronicles synoptic material and from foundational
studies of poetic parallelism, balanced against 825 randomly sampled negatives.
Through cosine-similarity regression, the model learns an embedding space in which
parallel verses cluster together and unrelated verses move apart. We evaluate
separation with distribution-based metrics, Wasserstein distance and the overlap
coefficient, across ten random seeds. MiqraBERT improves distributional separation
2.7-fold over the pre-trained baseline and reduces the ambiguous overlap region
from roughly 24\% to about 6\%. Narrative synoptic parallels reach a recall@10 of
87.1\%; poetic parallels remain difficult, below 9\%. This genre-dependent
asymmetry confines the model's reliable scope to narrative textual reuse.
MiqraBERT is publicly available at \url{https://huggingface.co/davidmsmiley/MiqraBERT}.
\end{abstract}

\noindent\textbf{Keywords:} Biblical Hebrew; Sentence-BERT; semantic textual
similarity; intertextuality; parallel detection; AlephBERT; transfer learning

\bigskip

\section{Introduction}
\label{sec:introduction}

Parallel texts and passages that reference others run throughout the Hebrew Bible (HB).\footnote{The literature on the nature of biblical parallelism is extensive. For the foundational arguments that parallelism involves coordinated variation across multiple linguistic dimensions rather than simple restatement, see Robert Alter, \textit{The Art of Biblical Poetry}, 2nd ed. (New York: Basic Books, 2011); Adele Berlin, \textit{The Dynamics of Biblical Parallelism}, rev. and exp. ed. (Grand Rapids: Eerdmans, 2008); James L. Kugel, \textit{The Idea of Biblical Poetry: Parallelism and Its History} (New Haven: Yale University Press, 1981). A fuller treatment of this theoretical background lies beyond the scope of the present study.} Large stretches of Chronicles parallel Samuel-Kings, individual oracles recur in variant forms across the prophetic collections, and poetic cola echo one another within and between books. Scholars have catalogued these correspondences for centuries, and the resulting reference works remain foundational for the field, yet the catalogues are incomplete. Each reflects its compilers' expertise and the limits of what any single reader can detect across a corpus of this size. Computational methods have been applied to the problem since the mid-2000s, but the dominant approaches still rest on lexical overlap. Some handle near-verbatim parallels adequately; almost all break down once correspondence involves paraphrase, lexical substitution, or syntactic restructuring. Neural network methods have begun to push past this ceiling in Classics, yet no model has been finetuned for verse-level semantic similarity in Biblical Hebrew (BH). This paper surveys those predecessors and presents MiqraBERT.

MiqraBERT is a regression-based Sentence-BERT finetuning of AlephBERT, a Modern Hebrew (MH) language model that has been adapted to detect parallel passages in the Hebrew Bible.\footnote{David M. Smiley, ``Intertextual Parallel Detection in Biblical Hebrew: A Transformer-Based Benchmark,'' \textit{arXiv:2506.24117} (2025): 7--8.} The training data comprises 1,650 labeled verse and half-verse pairs: 825 true parallels drawn from Chronicles synoptic material and four foundational studies of poetic parallelism, alongside 825 randomly sampled negatives.\footnote{David M. Smiley, ``T'OMIM: Tanakh Observable Matches of Intertextual Mimesis,'' \textit{Zenodo:19135731} (2026).} Through cosine similarity regression, MiqraBERT reshapes AlephBERT's embedding space so that parallel verses attract one another while non-parallel verses separate.

The results confirm that transfer learning from MH to BH is viable for parallel detection. MiqraBERT achieves a 2.7-fold improvement in distributional separation from its pre-trained AlephBERT at the optimal training configuration and reduces the ambiguous overlap region from 24\% to roughly 6\% of the embedding space. Narrative synoptic parallels reach a recall@10 of 87.1\%, placing the true parallel within the top ten candidates for nearly nine of every ten queries. Poetic parallelism proves resistant, with recall@10 below 9\%, exposing a genre-dependent asymmetry that constrains the model's current scope to narrative textual reuse.

\section{Related Work}
\label{sec:related-work}

Computational approaches to parallel passage detection in the Hebrew Bible have developed in two phases. Pre-neural network methods, which match surface-level features such as shared lexemes and string similarity, established the feasibility of automated detection. However, they encountered consistent limitations when textual reuse involved paraphrasing, lexical substitutions, or syntactic variation with little or no shared vocabulary. Neural network methods, developed more recently for Latin, Greek, and Hebrew corpora, have attempted to overcome this ceiling by learning distributional representations that encode semantic relationships beyond raw lexical overlap. The following review traces both phases, beginning with the pre-neural network approaches that define the problem our language model is set to address.

\subsection{Pre-Neural Network Computational Approaches in the Hebrew Bible}
\label{subsec:pre-neural-network-hb}

Van Peursen and Talstra conducted one of the earliest computational comparisons of synoptic passages on the Bible by analyzing 2 Kings 18--19 alongside its parallels in Isaiah 36--38 and 2 Chronicles 32.\footnote{Willem van Peursen and Eep Talstra, ``Computer-Assisted Analysis of Parallel Texts in the Bible: The Case of 2 Kings XVIII--XIX and Its Parallels in Isaiah and Chronicles,'' \textit{Vetus Testamentum} 57/1 (2007): 45--72.} Their method combined syntactical analysis with lexical correspondence matching, which were also supplemented by lexicons and frequency lists. Their work surfaced correspondences between Kings and Chronicles that traditional verse-aligned synopses, such as Bendavid and Endres,\footnote{Aba Bendavid, ed., \textit{Parallels in the Bible} [in Hebrew] (Carta, 2013); John Carol Endres et al., eds., \textit{Chronicles and Its Synoptic Parallels in Samuel, Kings, and Related Biblical Texts} (Liturgical Press, 1998).} have missed, since the Chronicler scattered and recombined source materials across verse boundaries. The approach demonstrated what careful computational alignment could achieve within a defined text set, but it required human intervention at each alignment step and was designed for known parallel passages rather than corpus-wide discovery on the HB.

Naaijer and Roorda developed an open-source algorithm that searched the entire HB for parallel passages by comparing text chunks using two complementary similarity measures. The first was SET, which computed the ratio of shared to total lexemes between passages. The second, LCS, applied Levenshtein distance at the string level.\footnote{Martijn Naaijer and Dirk Roorda, ``Parallel Texts in the Hebrew Bible, New Methods and Visualizations'' (2016), https://hal.science/hal-01283051v1.} The best results emerged from thresholds that were neither too permissive (producing chains of loosely connected ``cliques'') nor too strict (missing genuine parallels with lower lexical overlap). Using verse-level chunks, the tool recovered known parallels such as 2 Samuel 22 and Psalm 18, and contains a case study on 2 Kings 19--25, which confirmed parallels in Isaiah, Jeremiah, Chronicles, and other books. However, the study is wholly reliant on lexical overlap, which means that passages with structural or semantic correspondence without shared vocabulary fell outside the detectable range of both measures. For example, the tool itself struggled with the 2 Kgs 18:13--19:37 and 2 Chr 32:9--32 parallel, which Van Peursen and Talstra had analyzed. This section contains a significant difference in Chronicles where there is a lack of lexical dependency and falls below the algorithm's detection threshold.\footnote{Ibid., 7.}

Lobbezoo has written the most comprehensive systematic pre-neural network comparison of feature types for parallel detection in the Hebrew Bible.\footnote{Bert Lobbezoo, Computer-Based Recognition of Intertextuality within the Hebrew Bible (MA thesis, Delft University of Technology, 2015).} By building a ground truth dataset of cross-reference lists shared in the English Bible translations of the ESV, NIV, and NAS (10,788 verse pairs), he framed his approaches as a set of binary classification problems.\footnote{Ibid., 23.} Against this dataset, with an equal number of randomly sampled negative pairs and ten-fold cross-validation, Lobbezoo evaluated three feature families: 1) distance-based string metrics (Levenshtein, Jaro-Winkler, Dice's coefficient, and others, computed at character, word, n-gram, and word n-gram scales), 2) high-level grammatical features drawn from the WIVU/ETCBC morphosyntactic database,\footnote{Dirk Roorda et al., ``LAF-Fabric: A Data Analysis Tool for Linguistic Annotation Framework with an Application to the Hebrew Bible,'' \textit{arXiv:1410.0286} (2014). WIVU is an earlier version of the now BHSA dataset, see Willem van Peursen, Constantijn Sikkel, and Dirk Roorda, ``Hebrew Text Database ETCBC4b'' (Eep Talstra Centre for Bible and Computing, VU University Amsterdam, and Data Archiving and Networked Services, Royal Netherlands Academy of Arts and Sciences, 2015).} and 3) multi-scale distance features computed at sub-verse granularities (sentence, clause, and phrase levels).

The distance-based approach performed best, most likely because Hebrew lexemes typically consist of tri-consonantal roots where a typical word-level matching algorithm can detect that two words share a lexeme regardless of morphological inflection. Combining all string metrics with a logistic classifier yielded an AUC of 0.937,\footnote{Ibid., 42. AUC (Area Under the Receiver Operating Characteristic Curve) measures how well a classifier ranks positive examples above negative ones across all possible decision thresholds: a score of 1.0 indicates perfect discrimination, while 0.5 indicates performance no better than chance. For an accessible introduction, see Tom Fawcett, ``An Introduction to ROC Analysis,'' \textit{Pattern Recognition Letters} 27/8 (2006): 861--74. Lobbezoo reports performance as ``AUC error,'' an inverted convention in which lower values indicate better classification: 28--9. All AUC figures in this section have been converted to standard AUC (1 minus AUC error) for readability.} while the multi-scale approach produced comparable but slightly lower results, with clause-level distances reaching an AUC of 0.912.\footnote{Ibid., 62--3.} One caveat deserves attention; however, English translations of the biblical text slightly outperformed Hebrew lexemes in the distance-based experiments.\footnote{Ibid., 47.} This result may reflect bias in the ground truth dataset itself, since the cross-reference labels were compiled with translations that were aware of intertextual connections at the outset.

The grammatical approach tells a different story. Lobbezoo draws on the ETCBC database's 162 morphosyntactic features, covering part of speech, gender, number, person, verbal stem, verbal tense, phrase type, phrase function, clause type, and clause relation, among others. The best individual feature, the adverb part-of-speech tag, only achieved an AUC of 0.577.\footnote{Ibid., 50.} Of the remaining 161 features, 146 scored below 0.55, a range Lobbezoo himself characterizes as ``really close to random guessing''.\footnote{Ibid., 49.} Combining all 162 features with a logistic classifier raised the AUC only to 0.646. This arguably becomes Lobbezoo's most important contribution, namely, that no matter how richly encoded a dataset on the Hebrew Bible is, the high-level morphosyntax will not be the layer where the discriminatory signals for intertextual relationships reside.

\subsection{Pre-Neural Network Computational Approaches in Other Classical Semitic Texts}
\label{subsec:pre-neural-network-other}

Other works approach parallel detection at a much larger scale, such as Shmidman, Koppel, and Porat's algorithm for detecting textual reuse across Hebrew and Aramaic corpora.\footnote{Avi Shmidman, Moshe Koppel, and Ely Porat, ``Identification of Parallel Passages Across a Large Hebrew/Aramaic Corpus,'' \textit{Journal of Data Mining and Digital Humanities} (2018): 1388.} To overcome the computationally challenging task of targeting parallel passages of twenty words or more in the Babylonian Talmud, a 37-tractate compendium of rabbinic legal and exegetical discussion spanning approximately 1.8 million words, they reduced each word to its two least frequent letters.\footnote{An all-against-all comparison of twenty-word passages across the corpus would require approximately 1.6 trillion Levenshtein distance calculations. Shmidman, Koppel, and Porat estimated roughly twenty years of processing time on a single machine. See ibid., 3.} This normalized the orthographic variation that is pervasive in classical Hebrew and Aramaic texts, and, in effect, narrowed the search space enough for efficient indexing.\footnote{Ibid., 3--6. The most frequent Hebrew letters (\textit{yod}, \textit{aleph}, and \textit{vav}) happen to be the \textit{matres lectionis} and common prefixes, so retaining only the two rarest letters per word effectively strips orthographic noise.} The algorithm then matched skip-grams (four-word subsequences drawn from five-word windows) with allowances for gaps and interpolations of up to eight words in either passage.\footnote{Ibid., 5--7.} Processing the entire Talmud in eleven seconds on a single thread, the algorithm identified 4,602 parallel passage pairs with high precision on inspection.\footnote{Ibid., 7. Precision was assessed by manual inspection of results. Recall was estimated by comparison with exhaustive Levenshtein-based matching on a tractate-pair subset (\textit{Bava Batra} vs. \textit{Gittin}, 41 passages), where the algorithm recovered all matches plus five additional valid parallels that the edit-distance method had missed due to multi-word interpolations.} An auto-generated lexical substitution list, built iteratively from one-word discrepancies within the detected parallels, raised the total to 5,272 pairs in a second pass.\footnote{Ibid., 8. A third iteration yielded no further gain.}

The algorithm's efficiency, however, rests on that initial two-letter reduction, which ties detection to surface orthographic patterns. Even the substitution list addresses only synonyms that already appear in detected parallel contexts. Passages related by semantic correspondence without shared root vocabulary lie beyond what letter-frequency hashing can recover.

In a subsequent study, Shmidman refined the hashing approach for a harder problem: detecting biblical allusions in later Hebrew literature, both medieval and modern, where the borrowed material may be as short as two words and morphologically reworked to fit a new context.\footnote{Avi Shmidman, ``Automatic Identification of Biblical Citations and Allusions in Hebrew Texts,'' in \textit{Jewish Studies in the Digital Age} (De Gruyter, 2022), 336--8.} Short, shared phrases generate overwhelming numbers of false positives under naive matching, so a filtering principle was required.\footnote{To illustrate the scale of the false-positive problem: a naive two-word lexeme-pair search on just 198 words of Agnon returns 4,937 candidate matches across more than 3,400 biblical verses, a precision below 1\%, ibid., 338.} Shmidman's solution rests on the insight that the strongest predictor of a genuine allusion is the rarity of the matched phrase within the biblical text.\footnote{Ibid., 339.} The algorithm hashes sequences of two to five biblical words at three levels of representation (full vocalized form, vocalized form without prefix, and underlying lexeme), retaining only ``semi-unique'' combinations that occur in three or fewer verses.\footnote{Ibid., 341--4. Each word in the biblical text receives three representations: the full vocalized form, the base word with prefix removed, and the underlying lexeme. All permutations of these representations are computed for two- and three-word sequences, with skip-grams allowing for one-word gaps. Pruning reduces 27 million initial hashes to 2.5 million, storable in 80 MB of memory with near-instant lookup.}

Candidate matches in a target text are then scored by vocabulary rareness and penalized for interpolations, deletions, and morphological alterations.\footnote{Ibid., 345--6.} Evaluated on the first chapter of Abramowitch's \textit{The Fathers and the Sons} (1,929 words), where a domain expert identified 77 phrases constituting valid biblical allusions, the algorithm achieved 86\% recall and 81\% precision at the phrase level.\footnote{Ibid., 346--7. At the verse level, the algorithm achieved 83.5\% recall and 60\% precision, but the verse-level precision figure overstates the false-positive problem: inspection revealed that over half the misidentified verses were additional support verses for allusions the expert had already validated (346, n. 14).}

While this is the most refined lexical approach to Hebrew parallel detection to date, the morphological flexibility has a principled ceiling. Namely, an allusion must preserve at least one identifiable lexeme for the hash to match. Shmidman identifies paronomasia, where two words sound identical but derive from different lexemes, as with \texthebrew{שֵׁבֶט} (``staff'') and \texthebrew{שֶׁבֶת} (``idleness'') in Exodus 21:19, as such a boundary case.\footnote{Ibid., 347.}

These five pre-neural network studies converge on the same limitations. Whether operating on pre-selected test sets or searching an entire corpus, methods anchored in lexical overlap cannot capture the full dynamics of textual reuse that characterize much of the HB's transmission history. The problem, however, runs deeper than vocabulary matching alone. Each study produces binary or threshold-dependent decisions rather than continuous similarity scores, operates mostly at the sub-sentence level without representing a verse as a holistic semantic unit, and has no capacity to learn from labeled examples and generalize beyond its built-in rules. Neural network approaches to intertextual detection address these constraints.

\subsection{Neural Network Approaches to Parallel Detection}
\label{subsec:neural-network-approaches}

Neural networks which encode word/sentence embeddings, in particular, have begun to solve the issues found in lexical, pre-neural network approaches. Miller et al. integrated word embeddings into a classical alignment framework for Hebrew and Aramaic. Building on Brill et al.'s prior system,\footnote{Oran Brill, Moshe Koppel, and Avi Shmidman, ``FAST: Fast and Accurate Synoptic Texts,'' \textit{Digital Scholarship in the Humanities} 35/2 (2020): 254--264.} which paired the Needleman-Wunsch global alignment algorithm with word2vec embeddings, they introduced three modifications: 1) replacing Needleman-Wunsch with the Smith-Waterman local alignment algorithm to improve matching of full compositions with partial reuse fragments, 2) substituting fastText for word2vec to handle out-of-vocabulary terms through subword representations, and 3) adding Hebrew-specific enhancements to the distance function for orthographic variation, abbreviations, and multi-word alignment.\footnote{Hadar Miller, Tsvi Kuflik, and Moshe Lavee, ``Text Alignment in the Service of Text Reuse Detection,'' \textit{Applied Sciences} 15/6 (2025): 3395, 2--7.}

Evaluated on a ground truth dataset drawn from \textit{Leviticus Rabba} manuscript witnesses and Friedberg Project \textit{Talmud Bavli} variants (234 text pairs, 69,700 words), their best configuration achieved an F1 of 0.94, an eleven-percentage-point improvement over the Brill et al. baseline of 0.83. The gain was driven primarily by the switch to fastText, which alone accounted for a nine- to ten-point F1 increase, where the Smith-Waterman substitution contributed further by boosting recall on partial text reuse from 0.73 to 0.93.\footnote{See Table 1, ibid., 9--10.} Their work confirms that embedding-based features improve alignment quality for Hebrew and Aramaic, though the approach operates at the token-pair level, producing word-by-word alignments rather than sentence-level similarity scores.\footnote{Ibid., 8.}

Parallel work on Latin texts demonstrates that embedding representations can be expanded from the word-level to sentence-level. Burns et al. applied optimized word2vec embeddings to Valerius Flaccus's \textit{Argonautica} by evaluating against 945 known parallels catalogued from traditional commentaries, like Spaltenstein, Kleywegt, and Zissos.\footnote{The Tesserae search tool was considered state-of-the-art for Latin intertextual searching. Patrick J. Burns et al., ``Profiling of Intertextuality in Latin Literature Using Word Embeddings,'' in \textit{Proceedings of the 2021 Conference of the North American Chapter of the Association for Computational Linguistics: Human Language Technologies} (Association for Computational Linguistics, 2021), 4901--2.} Their model achieved 0.824 recall compared to 0.339 for the complete Tesserae result set and recovered intertexts sharing no lemmas at all, which is a category entirely unobtainable to the lexical methods reviewed above.\footnote{Ibid., 4902--3.}

Okuda's Latin SBERT applied a similar approach, but at the sentence level. By adapting the Sentence-BERT framework to Latin intertext discovery and providing the closest methodological precedent for the present study, Okuda rated the Lucan-Vergil benchmark, a dataset of 3,410 passage pairs from Vergil's \textit{Aeneid} and Lucan's \textit{Bellum Civile}, on a five-point intertext quality scale by annotators drawing on the commentary traditions.\footnote{Nozomu Okuda, ``Deep Learning for Intertext Discovery in Latin Epic: Latin SBERT and the Lucan-Vergil Benchmark'' (Ph.D. diss., State University of New York at Buffalo, 2022): 7.}

Okuda finetuned a Latin BERT encoder into a sentence-level similarity model. The finetuning procedure mirrors the Sentence-BERT architecture (see section 3.2). It, first, encodes paired passages separately, mean-pools them into sentence vectors, and compares them by cosine similarity, with human ratings converted to a 0-1 scale as training.\footnote{Ibid., 77--9.} On the binary meaningfulness task, which classifies parallels as meaningful (rated 3-5) or not (rated 1-2), Latin SBERT achieved a Matthews Correlation Coefficient (MCC) of 0.329, a 9.6\% increase over the previous best MCC of 0.3 achieved by Tesserae's lexical scores. On the finer-grained ratings prediction task, Latin SBERT more than doubled the optimized baseline, raising the MCC from 0.105 to 0.216.\footnote{See Table 2.3, ibid., 80--1.} These results in Latin focused case studies confirm that sentence-level representations can capture intertext relationships that word-level features and lexical scoring formulae cannot test.

McGovern et al. present the only neural network study targeting biblical text specifically. They apply contrastive learning to type-scene extraction across both Biblical Hebrew and Biblical Greek.\footnote{Hope McGovern et al., ``Detecting Narrative Patterns in Biblical Hebrew and Greek,'' in \textit{Proceedings of the 1st Workshop on Machine Learning for Ancient Languages} (Association for Computational Linguistics, 2024), 269--71.} By training on crowd-sourced biblical cross-references from openbible.info (17,899 Hebrew pairs and 45,297 Greek pairs), their architecture uses separate encoders for queries and candidates, instantiated from pre-trained BERT models and jointly trained with a triplet margin loss.\footnote{Ibid., 271--2.} They tested five base encoders spanning Ancient and Modern variants of both languages. Embible,\footnote{Niv Fono et al., ``Embible: Reconstruction of Ancient Hebrew and Aramaic Texts Using Transformers,'' \textit{Findings of the Association for Computational Linguistics} (Association for Computational Linguistics, 2024): 846--52.} an Ancient Hebrew encoder, and DictaBERT,\footnote{Shaltiel Shmidman, Avi Shmidman, and Moshe Koppel, ``DictaBERT: A State-of-the-Art BERT Suite for Modern Hebrew,'' \textit{arXiv:2308.16687} (13 October 2023).} a Modern Hebrew encoder converged to identical recall@10 of 0.08 after finetuning on the Leningrad Codex, up from 0.05 and 0.08 respectively in their pre-trained states.\footnote{See Table 3, McGovern, 272.} The best Greek model achieved a higher recall@10 of 0.12, likely reflecting the larger Greek training set.

All recall figures are computed at the chapter level, where the model must surface the correct chapter from among over 23,000 candidate verses for Hebrew or 31,000 for Greek. Manual error analysis showed that concrete type-scene queries with specific lexical anchors returned relevant candidates more reliably than abstract queries.\footnote{Ibid., 274--5.}

McGovern et al.'s findings bear directly on the present study. Mainly, the convergence of BH and MH encoders to the same post-finetuning performance indicates that for biblical similarity tasks, the choice of pre-trained base model may matter less than the quality of the finetuning signal. Additionally, the low absolute recall across both languages reflects the difficulty of the task at corpus scale. This present study takes this difficulty into consideration and establishes a distribution-based evaluation framework to address it.

\subsection{Hebrew-Specific Encoder Infrastructure}
\label{subsec:hebrew-encoder-infrastructure}

Since 2021 the encoder infrastructure for Hebrew NLP has expanded rapidly, producing domain-adapted models that span several registers of the language. AlephBERT, trained on 98.7 million Modern Hebrew sentences drawn from Common Crawl, Twitter, and Wikipedia, established state-of-the-art results on standard Hebrew benchmarks, including named entity recognition, sentiment analysis, and morphological disambiguation.\footnote{Amit Seker et al., ``AlephBERT: A Hebrew Large Pre-Trained Language Model to Start-off Your Hebrew NLP Application With,'' \textit{arXiv:2104.04052} (2021): 2--3.} BEREL followed with approximately 220 million words of rabbinic Hebrew from the Sefaria and Dicta libraries, designed as a general-purpose contextualized embedding model for downstream NLP tasks on Talmudic and midrashic texts.\footnote{Avi Shmidman et al., ``Introducing BEREL: BERT Embeddings for Rabbinic-Encoded Language,'' \textit{arXiv:2208.01875} (2022): 1--4.} MsBERT extended coverage to Hebrew manuscripts and was trained on over 67 million words spanning the third through thirteenth centuries for the specific task of reconstructing lacunae in damaged manuscripts.\footnote{Avi Shmidman et al., ``MsBERT: A New Model for the Reconstruction of Lacunae in Hebrew Manuscripts,'' in \textit{Proceedings of the 1st Workshop on Machine Learning for Ancient Languages (ML4AL 2024)} (ACL, 2024): 13--18.} Each of these models was designed for a different downstream application and a different temporal stratum of the language. None was finetuned for semantic similarity detection between different passages, and none targets the Hebrew Bible as its primary domain.

Among these candidates, AlephBERT holds a particular advantage for Biblical Hebrew tasks despite its Modern Hebrew training data. The triconsonantal root system that structures Semitic word formation persists across both registers, as do the \textit{binyan} verbal conjugation patterns that govern derivational morphology.\footnote{Edit Doron, ``The Biblical Sources of Modern Hebrew Syntax,'' in \textit{Language Contact, Continuity and Change in the Genesis of Modern Hebrew} (Amsterdam: John Benjamins, 2019), 222--56. Ora Rodrigue Schwarzwald, ``Morphology: Modern Hebrew,'' \textit{Encyclopedia of Hebrew Language and Linguistics Online} (Brill, 2013).} AlephBERT also outperformed multilingual alternatives on a direct benchmark of BH verse embeddings, achieving a Wasserstein Distance (WD) between parallel and non-parallel similarity distributions roughly three times larger than E5, the strongest multilingual model tested.\footnote{David M. Smiley, ``Intertextual Parallel Detection in Biblical Hebrew: A Transformer-Based Benchmark,'' \textit{arXiv:2506.24117} (2025): 7--8.} A wider distributional gap means more geometric room for finetuning to separate the two classes, a property that becomes important when the training signal is small.

McGovern et al.'s findings on biblical type-scene detection reinforce this point from a different angle. When they finetuned Embible and DictaBERT on crowd-sourced biblical cross-references, both models converged to an identical recall@10. The pre-trained register of the base model, whether ancient or modern, made no measurable difference once a biblical finetuning signal was applied. What mattered was the quality and relevance of the training pairs. This convergence suggests that the choice of Hebrew encoder for biblical retrieval tasks is less about matching the historical period of the target text than about selecting a model whose internal representations are amenable to the finetuning pipeline, of which AlephBERT's benchmark performance and relatively compact architecture give it an edge on other Hebrew language models.

What remains absent from this growing ecosystem is a model finetuned on curated parallel passages from the HB for the specific task of verse-level semantic similarity. The encoders surveyed above were optimized for token-level predictions (AlephBERT), passage completion (BEREL, MsBERT), or contrastive type-scene retrieval (Embible, DictaBERT), but not for producing continuous similarity scores between verse pairs. Filling this gap, however, is not simply a matter of applying any finetuning procedure to the best available base model. Pre-trained transformers carry a geometric distortion in their embedding spaces that compromises similarity measurement before any task-specific training begins. Adapting the Hebrew encoder infrastructure to parallel detection requires first understanding, and then correcting, this structural problem.

\section{Data and Methodology}
\label{sec:data-and-methodology}

\subsection{The Anisotropy Problem: Why Pre-Trained Models Fail}
\label{subsec:the-anisotropy-problem-why-pre-trained-models-fail}

Pre-trained transformer models suffer from a fundamental geometric flaw that compromises their utility for semantic similarity tasks. Specifically, before being finetuned, sentence embeddings are distorted in one of two ways. They either cluster into narrow cones within high-dimensional space or drift toward dominant directions which are inconsistent with a word's semantics. This phenomenon is called anisotropy, which refers to a side effect of word representation during model training that results in an embedding distribution that does not uniformly fill the embedding space.\footnote{Jun Gao et al., "Representation Degeneration Problem in Training Natural Language Generation Models," in \textit{International Conference on Learning Representations }(2019): 2–6. Kawin Ethayarajh, "How Contextual are Contextualized Word Representations? Comparing the Geometry of BERT, ELMo, and GPT-2 Embeddings," in \textit{Proceedings of the 2019 Conference on Empirical Methods in Natural Language Processing and the 9th International Joint Conference on Natural Language Processing (EMNLP-IJCNLP)} (2019): 57–58.} Since pre-trained models are being trained for general text representation, their modeling objectives produce an embedding that is not optimized for text similarity tasks.

From the earlier study which benchmarked pre-trained language models on BH, the most relevant effects of anisotropy manifest in two ways.\footnote{See especially the cosine similarity score distribution graphs in the Appendix of David M. Smiley, "Intertextual Parallel Detection in Biblical Hebrew: A Transformer-Based Benchmark," \textit{arXiv:2506.24117} (2025): 10.} First, high-frequency words cluster around the origin of the embedding space, despite any semantic information gained from the modeling process. Second, low-frequency words become disparate data points and find themselves either in "holes" or scattered into isolated geometric coordinates where their semantic meaning is poorly defined.\footnote{Bohan Li et al., "On the Sentence Embeddings from Pre-trained Language Models," in \textit{Proceedings of the 2020 Conference on Empirical Methods in Natural Language Processing (EMNLP)} (2020), 9121–22.}

In the case of AlephBERT, the model was trained on a range of Hebrew texts, and with just under 23,000 verses, even if every biblical passage makes its way into the training set multiple times over, they will concentrate into their own group for two reasons. First, by being a different form of Hebrew that is distinct from say the innumerable shitposts that are undoubtedly embedded in the model.\footnote{"Shitposting is the act of throwing out huge amounts of content, most of it ironic, low-quality trolling, for the purpose of provoking an emotional reaction in less Internet-savvy viewers. The ultimate goal is to derail productive discussion and distract readers." Robert Evans, "Shitposting, Inspirational Terrorism, and the Christchurch Mosque Massacre," Bellingcat, March 15, 2019, accessed December 8, 2025, https://www.bellingcat.com/news/rest-of-world/2019/03/15/shitposting-inspirational-terrorism-and-the-christchurch-mosque-massacre/. See also, Peter Woods, "Shitposting as Public Pedagogy,"\textit{ Curriculum Inquiry} 53/4 (2023): 359–80.} Second, the self-attention mechanism of a transformer inherently causes a concentration in low-frequency words.\footnote{Nathan Godey et al., "Anisotropy Is Inherent to Self-Attention in Transformers," in\textit{ Proceedings of the 18th Conference of the European Chapter of the Association for Computational Linguistics, Volume 1: Long Papers} (2024): 36–37. Daniel Biś et al., "Too Much in Common: Shifting of Embeddings in Transformer Language Models and its Implications," in\textit{ Proceedings of the 2021 Conference of the North American Chapter of the Association for Computational Linguistics: Human Language Technologies }(2021): 5121–22.} Upon updating the representations of rarer words there is a tendency for a "drift" to occur in a consistent direction because of the lower probability of seeing a BH verse when sampling from the originally trained Hebrew corpus.\footnote{For more on how the softmax function normalizes embeddings by converting them to next-token predictions, see Ashish Vaswani et al., "Attention Is All You Need," in\textit{ Proceedings of the 31st Conference on Neural Information Processing Systems (NIPS) }(2017): 6005–6.}

\begin{figure}[htbp]
  \centering
  \includegraphics[width=0.85\textwidth]{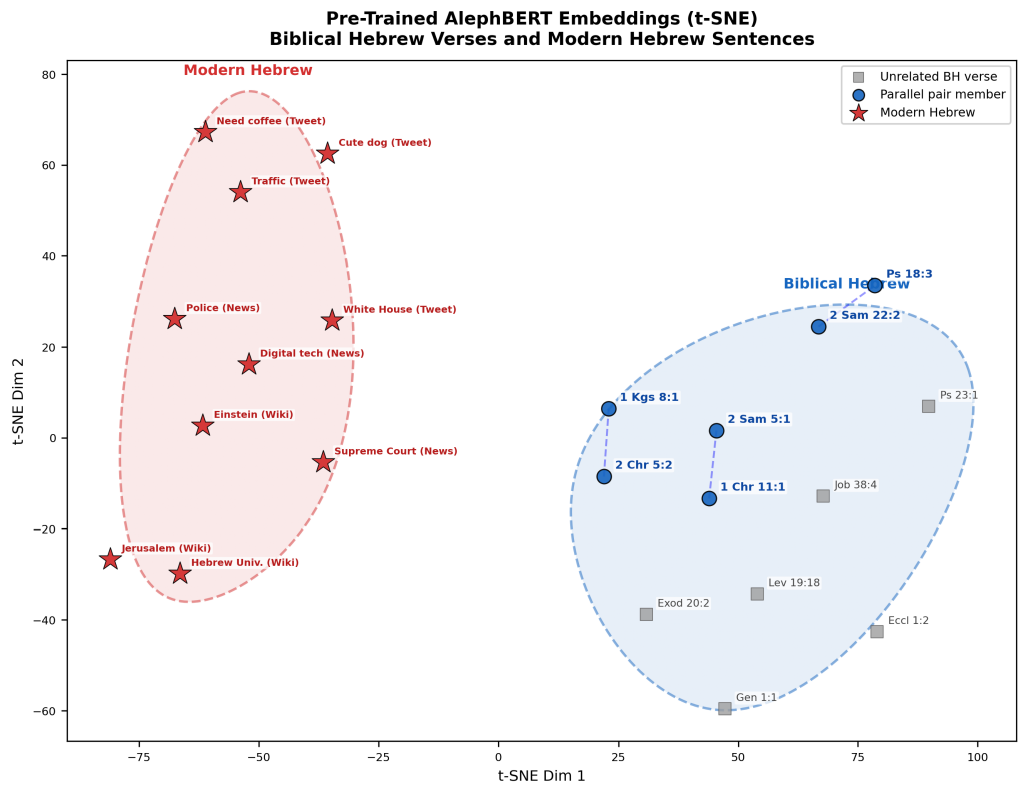}
  \caption{t-SNE projection of pre-trained AlephBERT embeddings,  Biblical Hebrew vs. Modern Hebrew}
  \label{fig:fig1_alephbert_tsne_pretrained}
\end{figure}

Figure 3.1 confirms this concentration, as the t-SNE projection of AlephBERT's sentence embeddings shows two distinct clusters: Biblical Hebrew verses occupy one region of the reduced space, while Modern Hebrew sentences drawn from AlephBERT's own training data (Wikipedia entries, news articles, and tweets) form another. The model separates BH from MH probably on the basis of register alone. However, the real problem lies inside the BH cluster. Three known synoptic parallel pairs (2 Sam 5:1//1 Chr 11:1; 2 Sam 22:2 // Ps 18:3; 1 Kgs 8:1 // 2 Chr 5:2) are fixed inside the "Biblical Hebrew" region with no unique spatial proximity to their genuine counterparts. The pre-trained, untuned model treats every biblical verse as a variation on a single theme: it is Biblical Hebrew, and that is all the embedding encodes. This is the narrow-cone problem in concrete form, where an entire low-frequency register collapses into one geometric neighborhood. Contrastive finetuning reshapes this space so that genuine parallels attract one another while unrelated passages separate. That transformation is the subject of the next section.

Let us consider two verses which share no thematic, generic, or theological relationship: 2 Sam 5:1 and Job 38:4. In the former, the tribes of Israel pledge their allegiance to David after the death of Saul, while the latter presents YHWH interrogating Job about where he was when the foundations of the earth were laid. No scholar would identify any parallel or intertextual relationship between the two; however, E5 (the best performing multilingual model) assigns this non-pair a cosine similarity of 0.87, while giving a score of 0.97 to 2 Sam 5:1 and its genuine synoptic parallel in 1 Chr 11:1. AlephBERT performs a bit better and only gives 2 Sam 5:1 and Job 38:4 a cosine similarity of 0.52, which is still too high given there is no lexical or semantic overlap at all.

As for the language-agnostic and multilingual models, this problem is exacerbated simply by attempting to contain the Hebrew language, whether Modern or biblical, into one cluster of a total potential embedding space that contains many other languages and dialects. These are precisely the conditions that drive semantic embedding collapse in pre-trained models. Contrastive learning addresses this geometric problem directly, reshaping embedding spaces in a way that parallel passages cluster together and separate themselves from non-parallel passages.\footnote{Tianyu Gao, Xingcheng Yao, and Danqi Chen, "SimCSE: Simple Contrastive Learning of Sentence Embeddings," in \textit{Proceedings of the 2021 Conference on Empirical Methods in Natural Language Processing} (2021): 6898. Lingling Xu et al., "Contrastive Learning Models for Sentence Representations," \textit{ACM Transactions on Intelligent Systems and Technology }14/4 (2023): 7–14.}

\subsection{Finetuning with Sentence-BERT}
\label{subsec:finetuning-with-sentencebert}

Addressing anisotropy requires an appropriate finetuning approach which, in our case, also modifies model weights for semantic similarity tasks. AlephBERT was selected as the primary finetuning candidate for three primary reasons: 1) linguistic proximity to BH, 2) superior performance in the benchmark study, and 3) unlike multilingual models like E5, it is a fairly small and, thus, computationally efficient language model.

First, language-specific pre-training provides advantages for BH that multilingual models cannot replicate. MH morphology remains fundamentally rooted in the biblical and classical versions of the language and preserves the triconsonantal root system, as well as the binyan verbal conjugation patterns that structure Semitic formation.\footnote{Edit Doron, "The Biblical Sources of Modern Hebrew Syntax," in\textit{ Language Contact, Continuity and Change in the Genesis of Modern Hebrew} (ed. Edit Doron et al.; LT 256; Amsterdam: John Benjamins, 2019), 222–56. Yael Reshef, \textit{Historical Continuity in the Emergence of Modern Hebrew }(Lanham, MD: Lexington Books, 2019).} While lexical continuity proves more difficult to quantify, estimates of vocabulary deriving from biblical and other classical sources range from 22\% to 65\% depending on classification methodology.\footnote{Ora Rodrigue Schwarzwald, "Lexicon: Modern Hebrew,"\textit{ Encyclopedia of Hebrew Language and Linguistics Online} (Brill, 2013). See also Schwarzwald, "The Components of the Hebrew Lexicon: The Influence of Hebrew Classical Sources, Jewish Languages and Other Foreign Languages on Modern Hebrew,"\textit{ Hebrew Linguistics} 39 (1995): 79–90; Bat-Zion Yamini, "The Revival of Ancient Hebrew Words With the Revival of Israel,"\textit{ Sociology Study} 9/4 (2019): 160; Even-Shoshan, \textit{The New Dictionary }[in Hebrew] (Jerusalem: Kirjat Sepher, 1970), 3062.} However, the morphological structure of MH exhibits substantial persistence with BH.\footnote{"Modern Hebrew morphology is based on classical, mainly Biblical Hebrew, albeit with many changes, some related to word structure (the lexicon), others due to morphophonemic and morphosyntactic factors," in Schwarzwald, "Morphology: Modern Hebrew," \textit{Encyclopedia of Hebrew Language and Linguistics Online} (Brill, 2013).} This continuity positions AlephBERT as a plausible candidate for transfer learning despite any temporal distance separating its MH training corpora from the BH target texts.

This relative advantage was confirmed in the benchmark study. All pre-trained models exhibited anisotropy-induced compression, but AlephBERT achieved a WD of 0.276 compared to 0.081 for E5. This threefold difference indicates a geometric structure more amenable to finetuning, as its main task will be weeding out the false positives. AlephBERT's Hebrew-specific representations provide a foundation that our similarity regression can reshape. Rather than requiring the model to "learn" BH morphology as a part of the downstream task, it can immediately leverage these features and focus its computational resources on adapting to BH parallel detection.

The alternative to finetuning would be to train a domain-specific encoder on the Hebrew Bible itself, something that would not be possible using a transformer-based model. The corpus is simply too small for the attention mechanism to learn the appropriate weights. The Hebrew Bible contains 426,590 tokens (according to BHSA word nodes), which falls short of the 22 million minimum threshold by a factor of roughly fifty.\footnote{Jared Kaplan et al. demonstrated that transformer performance follows a power-law relationship with dataset size and their smallest functional training set contained approximately 22 million tokens, "Scaling Laws for Neural Language Models,"\textit{ arXiv:2001.08361} (2020): 7, 11.}

\begin{figure}[htbp]
  \centering
  \includegraphics[width=\textwidth]{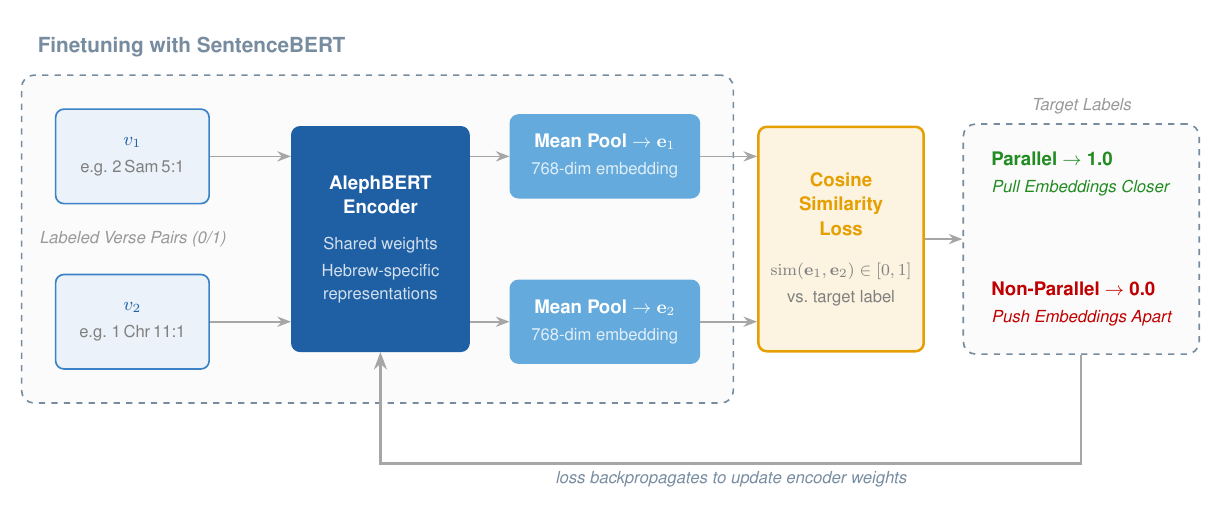}
  \caption{Sentence-BERT finetuning pipeline. Both input verses pass through a shared AlephBERT encoder; mean pooling produces fixed-length embeddings; cosine similarity is compared against the target label via regression loss.}
  \label{fig:fig2_sbert_finetuning_pipeline}
\end{figure}

With our chosen pre-trained model, an appropriate finetuning framework must also address the anisotropy problem directly. Sentence-BERT modifies pre-trained BERT models for semantic similarity through siamese BERT-networks.\footnote{Nils Reimers and Iryna Gurevych, "Sentence-BERT: Sentence Embeddings Using Siamese BERT-Networks," in \textit{Proceedings of the 2019 Conference on Empirical Methods in Natural Language Processing }(Hong Kong, China: Association for Computational Linguistics, 2019), 3980–90.} The implementation uses a single encoder with shared weights (see Figure 3.2): both input verses pass through identical AlephBERT models. For each verse pair ($v_1$, $v_2$), verses are encoded independently, and mean pooling converts token-level representations to verse-length vectors of 768 dimensions $(e_1, e_2)$. Cosine similarity between embeddings is then compared against target labels (parallel $=1$, non-parallel $=0$). While the labels themselves are binary, the predicted similarity is understood to be continuous.

\subsection{Training Objective and Loss Function}
\label{subsec:training-objective-and-loss-function}

In order to account for a continuum of similarity and degrees of semantic overlap, a non-binary classification objective is needed for our training pairs. If a classification objective is chosen, the model will train to predict discrete, delineated labels.\footnote{For more on binary classification based methods see Qian Li et al., "A Survey on Text Classification: From Traditional to Deep Learning," \textit{ACM Transactions on Intelligent Systems and Technology} 13/2 (2022).} This will collapse the gradation needed for nuance and enforce binary "parallel/non-parallel" predictions.

Regression-based finetuning, however, addresses this limitation by training the model to predict continuous similarity scores (\textit{toward} 1.0 for a parallel, \textit{toward} 0.0 for non-parallel) rather than discrete categories.\footnote{Reimers and Gurevych, "Sentence-BERT," 3985 demonstrates the regression objective with graded similarity targets (0 to 5, normalized). Our implementation applies mean squared error to binary targets (0 and 1), but the continuous prediction remains essential. The model outputs similarity scores across the full range of 0 and 1, rather than collapsing choices to a 0.5 decision threshold that applies a binary label. This preserves the gradient of similarity that will allow us to examine not only whether texts are parallel but the degree to which they are similar. Thus, preserving the interpretation of similarity ultimately with the reader.} The loss function then minimizes the mean squared error between predicted similarity and target similarity.\footnote{The cosine similarity regression loss is defined as $\mathcal{L} = (s - t)^2$, where $s = \cos(\mathbf{e}_1, \mathbf{e}_2)$ is the cosine similarity between the two verse embeddings and $t \in \{0, 1\}$ is the target label.} Through backpropagation, the encoder weights are recalibrated from the cosine similarity difference with its target label. This training dynamic, illustrated in Figure 3.3, produces a clear distributional separation, as will be shown later in the results.

\begin{figure}[htbp]
  \centering
  \includegraphics[width=0.9\textwidth]{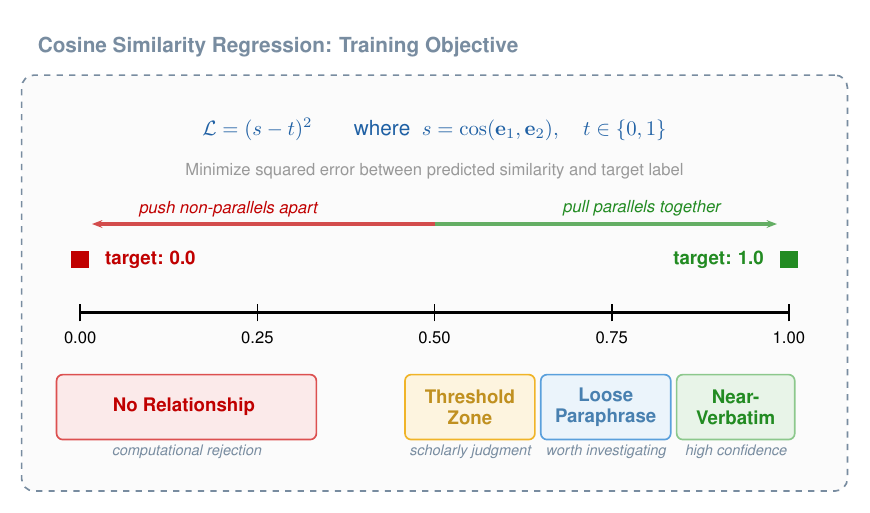}
  \caption{Cosine similarity regression training objective. The Siamese network passes both verses through the shared encoder, computes cosine similarity between the resulting embeddings, and minimizes the squared error against the target label.}
  \label{fig:fig3_training_objective}
\end{figure}

\subsection{Dataset Construction}
\label{subsec:dataset-construction}

The training data consists of 1,650 half-verse and verse pairs (825 true parallels, 825 non-parallels) constructed to represent both literary and non-literary genres of biblical parallelism.\footnote{For a full list of parallels see David M. Smiley, ``T'OMIM: Tanakh Observable Matches of Intertextual Mimesis''. \textit{Zenodo:19135731} (2026).} By capturing parallelism across multiple genres the aim is to produce a model capable of detecting textual relationships regardless of literary context. The sources for true parallels are provided below with each fitting the criteria of locality and linguistic correspondence established by Bothwell et al., which was used in their attempts to detect rhetorical parallelism in Latin text.\footnote{Stephen Bothwell et al., "Introducing Rhetorical Parallelism Detection: A New Task with Datasets, Metrics, and Baselines," in \textit{Proceedings of the 2023 Conference on Empirical Methods in Natural Language Processing} (2023), 5008.}

There are 556 narrative synoptic parallels that derive from the synoptic and shared materials of Chronicles and Samuel-Kings.\footnote{John Carol Endres et al., eds., \textit{Chronicles and Its Synoptic Parallels in Samuel, Kings, and Related Biblical Texts} (Collegeville, MN: Liturgical Press, 1998).} For a model architecture that is focused on capturing semantic meaning, and not merely lexical overlap, this corpus contains a rich spectrum of differences between the parallel material.\footnote{On other character-, phrase-, and corpus-based similarity measures see Jiapeng Wang and Yihong Dong, "Measurement of Text Similarity: A Survey," \textit{Information} 11/9 (2020): 5–8.} As Kalimi notes in \textit{The Reshaping of Ancient Israelite History in Chronicles},\footnote{Isaac Kalimi,\textit{ The Reshaping of Ancient Israelite History in Chronicles }(University Park, PA: Penn State University Press, 2012).} even corresponding verses in the later Chronicles material range from (near-)verbatim quotations to moderate/heavily redacted paraphrases. The scope will allow our model to maintain a balance of locating texts that contain lexical overlap while also remaining flexible to generalize well to semantically similar, yet non-verbatim texts.

The poetic parallels (269 pairs) were drawn from four foundational studies in BH parallelism.\footnote{Adele Berlin, \textit{The Dynamics of Biblical Parallelism}, (rev. and exp. ed.; Grand Rapids, MI: Eerdmans, 2008); Joannes Petrus Fokkelman, \textit{Reading Biblical Poetry: An Introductory Guide}, trans. Ineke Smit (Louisville, KY: Westminster John Knox Press, 2001); James L. Kugel, \textit{The Idea of Biblical Poetry: Parallelism and Its History} (New Haven: Yale University Press, 1981); David Toshio Tsumura, \textit{Vertical Grammar of Parallelism in Biblical Hebrew} (Atlanta: SBL Press, 2023).} Poetic parallelism provides some unique features that are missing in the narrative synoptic material. They often operate at the half/sub-verse level and, instead of containing a great deal of lexical overlap, they capture semantic pairs, as well as structural similarities. Training on both narrative and poetic examples ensures that the model learns parallelism as a multidimensional phenomenon that encompasses lexical, structural, and morphological relationships rather than a simple edit distance metric.\footnote{For more on edit distance see Vladimir I. Levenshtein, "Binary Codes Capable of Correcting Deletions, Insertions, and Reversals," \textit{Soviet Physics Doklady }10 (1965): 707–710; Daniel Jurafsky and James H. Martin, \textit{Speech and Language Processing }(3rd ed.; online manuscript, 2026), 30–35.}

The imbalance between narrative and poetic examples (556 vs. 269 pairs) reflects source availability rather than a deliberate design feature. Comprehensive synoptic references in Chronicles provide systematic coverage of narrative parallelism, while poetic examples require manual extraction from scholarship that lacks exhaustive catalogues. This asymmetry will carry methodological consequences, as will be shown in the latter sections of this paper, and is acknowledged as a point of future refinement.

Negative parallels (825 pairs) were generated through random negative sampling (RNS) rather than soft negative parallels with antithetical parallelism.\footnote{On soft negative sampling for contrastive learning see Hao Wang and Yong Dou, "SNCSE: Contrastive Learning for Unsupervised Sentence Embedding with Soft Negative Samples," in \textit{Advanced Intelligent Computing Technology and Applications }(ed. De-Shuang Huang et al.; Singapore: Springer Nature Singapore, 2023), 419–31.} Excluding the positive parallels listed above, random verse pairs were selected without replacement from the entire HB. RNS provides a more stable and computationally efficient baseline for the model and will provide examples of non-parallels so that our model avoids the numerous false positives found in the pre-trained benchmarked models.\footnote{Lanling Xu et al., "Negative Sampling for Contrastive Representation Learning: A Review," \textit{arXiv:2206.00212} (2022): 3–8.} While more sophisticated, dynamic or adversarial sampling strategies are available, RNS is the most prevalent method for constructing negative training pairs in contrastive learning tasks and serves as the standard baseline against which other sampling strategies are evaluated.\footnote{Zhen Yang et al., "Does Negative Sampling Matter? A Review with Insights into Its Theory and Applications," \textit{IEEE Transactions on Pattern Analysis and Machine Intelligence} 46/8 (2024): 5698; Qitao Tan et al. as an example of using random sampling as foundational method to test better methods: "most contrastive learning-based models leverage random strategy to construct negative pairs, and such pairs are at risk of being uninformative," "An Effective Negative Sampling Approach for Contrastive Learning of Sentence Embedding," \textit{Machine Learning} 112 (2023): 4841.}

\subsection{Data Allocation and Multi-Seed Validation}
\label{subsec:data-allocation-and-multi-seed-validation}

Optimal allocation between training, testing, and validation required empirical investigation across the 1,650 labeled pairs. Seven allocation strategies were evaluated, ranging from 50\%-25\%-25\% to 90\%-5\%-5\% training-testing-validation splits. All configurations maintained stratification across parallel/non-parallel labels and narrative/poetic genres to ensure balanced training-testing-validation sets.

\begin{table}[htbp]
  \centering
  \caption{Training-Validation-Test Allocation Configurations}
  \label{tab:table-35}
  \begin{tabular}{lccc}
    \toprule
    Configuration (\%) & Training (\textit{n}) & Validation (\textit{n}) & Test (\textit{n}) \\
    \midrule
    50-25-25 & 825 & 412 & 413 \\
    60-20-20 & 990 & 330 & 330 \\
    70-15-15 & 1,155 & 247 & 248 \\
    75-12.5-12.5 & 1,237 & 206 & 207 \\
    80-10-10 & 1,320 & 165 & 165 \\
    85-7.5-7.5 & 1,402 & 124 & 124 \\
    90-5-5 & 1,485 & 82 & 83 \\
    \bottomrule
  \end{tabular}
\end{table}

Each configuration underwent evaluation using ten different random seeds to assess performance stability across different data splits. Stratified splitting ensures balanced representation of label combinations in all partitions. Models were trained independently for each seed-configuration combination, producing seventy total models.\footnote{Models were trained for 2 epochs, and preliminary experiments showed validation performance plateauing after the second epoch with additional training exhibiting early signs of overfitting. Training the 70\%-15\%-15\% configuration completed in approximately 36 seconds on a T4 GPU in a Google Colab environment. Once finetuned, the model encodes the full Hebrew Bible ({\raise.17ex\hbox{$\scriptstyle\sim$}}23,000 verses) in under 2 minutes on said GPU.}

Multi-seed validation addresses critical limitations of single-split evaluation. Finetuning pre-trained transformer models on small datasets can create model instability and yield misleading performance estimates.\footnote{Marius Mosbach et al., "On the Stability of Fine-tuning BERT: Misconceptions, Explanations, and Strong Baselines," in \textit{Proceedings of the 9th International Conference on Learning Representations (ICLR)} (Open Review, 2021). The authors found that finetuning BERT models over 25 random seeds produced significant variance.} Dodge et al. showed that test set performance scores alone are insufficient for drawing accurate conclusions about model comparisons, with some finetuning results varying as much as 7\% accuracy depending on random seeds.\footnote{Jesse Dodge et al., "Fine-Tuning Pretrained Language Models: Weight Initializations, Data Orders, and Early Stopping," \textit{arXiv:2002.06305} (2020): 2–3.} When datasets are small, particular random splits risk yielding misleading performance metrics.

Therefore, a model might appear strong simply because of chance with "easier" testing pairs. By quantifying this potential sensitivity to testing sets across multiple seeds, the performance variance will reveal more stable results for each split configuration. Thus, statistical testing across matched seeds enables rigorous comparisons that would be misleading with single-split designs.

\subsection{Evaluation Metrics}
\label{subsec:evaluation-metrics}

Two complementary metrics were employed to assess model performance: WD (as was noted in an earlier study in section 2.4) and the Overlap Coefficient (OVL). Distribution-based metrics quantify separation between similarity score distributions for parallel versus non-parallel pairs without requiring threshold selection, as was shown in the benchmark study. WD measures the minimum "work" required to transform the parallel distribution into the non-parallel distribution with higher values indicating better separation.\footnote{Wasserstein Distance, also known as Earth Mover's Distance, quantifies the cost of optimally transporting probability mass from one distribution to another. For statistical testing applications in comparing distributions, see Aaditya Ramdas et al., "On Wasserstein Two-Sample Testing and Related Families of Nonparametric Tests," \textit{Entropy} 19/2 (2017): 47.}

OVL represents the proportion of probability mass where distributions intersect and is computed via kernel density estimation. Values near zero indicate cleaner separation.\footnote{For the nonparametric kernel density estimators used here, see Friedrich Schmid and Axel Schmidt, "Nonparametric Estimation of the Coefficient of Overlapping—Theory and Empirical Application," \textit{Computational Statistics \& Data Analysis }50 (2006): 1583–1596.} Both distribution metrics serve as primary evaluation criteria because they measure representation quality independent of the threshold selection required by classification metrics, like F1. A model with high WD and low OVL organizes its embedding space such that parallel and non-parallel texts occupy distinct regions. Bootstrap confidence intervals (1,000 iterations, percentile method) quantify sampling uncertainty and for multi-seed experiments, standard deviation across seeds captures training instability.\footnote{Bradley Efron and Robert J. Tibshirani, \textit{An Introduction to the Bootstrap} (New York: Chapman \& Hall, 1993), 168–177.}

\section{Results of MiqraBERT}
\label{sec:results-of-miqrabert}

\subsection{Finetuning Performance}
\label{subsec:finetuning-performance}

Five-fold cross-validation across the ten random seeds confirmed that similarity regression finetuning transforms AlephBERT into an effective BH parallel detector. Pre-trained AlephBERT achieved an overall WD of 0.276 between parallel and non-parallel cosine similarity distributions, with parallel pairs averaging 0.816 and non-parallel pairs of 0.540. OVL stood at 0.240. This means that nearly a quarter of the embedding space for all of our pairs remained ambiguous between the two classes.

Finetuning produced improvements across all split configurations. Table 4.1 below reports confidence intervals computed from ten independent training runs with different random seeds.\footnote{The multi-seed protocol follows best practice for evaluating finetuned language models. See Dodge et al., "Fine-Tuning Pretrained Language Models," 1–5; See also Nils Reimers and Iryna Gurevych, "Reporting Score Distributions Makes a Difference: Performance Study of LSTM-networks for Sequence Tagging," in \textit{Proceedings of the 2017 Conference on Empirical Methods in Natural Language Processing} (Association for Computational Linguistics, 2017): 338–41.} For each configuration, we calculated the average score and then determined the 95\% confidence interval (CI) that the true average falls in, using the standard statistical formula for small samples by multiplying the standard error by the appropriate correction factor for ten observations.\footnote{That is, \textit{mean} ± \textit{t} × (\textit{SD}/√\textit{n}), where\textit{t} is drawn from the Student t-distribution with\textit{ n − 1} degrees of freedom. See William S. Gosset [Student], "The Probable Error of a Mean," \textit{Biometrika} 6 (1908): 1–25.} For example, the 70\%-15\%-15\% configuration achieved a mean WD of 0.751, with the 95\% CI ranging from 0.736 to 0.766. This represents a 2.7-fold improvement over the pre-trained baseline.\footnote{A non-neural baseline confirms that this improvement reflects more than shared vocabulary. TF-IDF cosine similarity, fitted on the full 22,946-verse Hebrew Bible corpus, achieves a WD of only 0.448 under the best tokenization (character 3-5 grams), which is roughly 60\% of MiqraBERT's 0.751. The word-level configuration fares worse still (WD = 0.406), in part because frequency filtering reduces the usable Hebrew vocabulary to just 103 terms. Biblical Hebrew's morphological richness and abundance of hapax legomena leave almost nothing in the mid-frequency range that TF-IDF requires. The gap between the two models is widest on narrative parallels, where MiqraBERT achieves a mean pairwise similarity of 0.884 against TF-IDF's 0.675, despite narrative parallels sharing substantial vocabulary with their source texts.} After finetuning, the model pushes parallel and non-parallel verse pairs much further apart in its internal embedding space. OVL between these two distributions fell by roughly 75\%, dropping to just 0.061. Therefore, only about 6\% of the representative space remains ambiguous as the region where the model cannot clearly distinguish a genuine parallel from a non-parallel pair, as opposed to 24\% by pre-trained AlephBERT.

\begin{table}[htbp]
  \centering
  \caption{Multi-Seed Performance Across Split Configurations}
  \label{tab:table-41}
  \begin{tabular}{llll}
    \toprule
    Configuration & Training Pairs & WD [95\% CI] & OVL [95\% CI] \\
    \midrule
    Pre-Trained & None & 0.276 & 0.240 \\
    50-25-25 & 825 & 0.719 [0.697, 0.741] & 0.060 [0.043, 0.077] \\
    60-20-20 & 990 & 0.744 [0.735, 0.752] & 0.063 [0.050, 0.075] \\
    70-15-15 & 1,155 & 0.751 [0.736, 0.766] & 0.061 [0.049, 0.073] \\
    75-12.5-12.5 & 1,237 & 0.753 [0.739, 0.767] & 0.064 [0.048, 0.081] \\
    80-10-10 & 1,320 & 0.765 [0.747, 0.782] & 0.062 [0.048, 0.077] \\
    85-7.5-7.5 & 1,402 & 0.757 [0.740, 0.775] & 0.077 [0.056, 0.097] \\
    90-5-5 & 1,485 & 0.776 [0.753, 0.799] & 0.050 [0.027, 0.073] \\
    \bottomrule
  \end{tabular}
\end{table}

\begin{figure}[htbp]
  \centering
  \begin{minipage}[t]{0.49\textwidth}
    \centering
    \includegraphics[width=\linewidth]{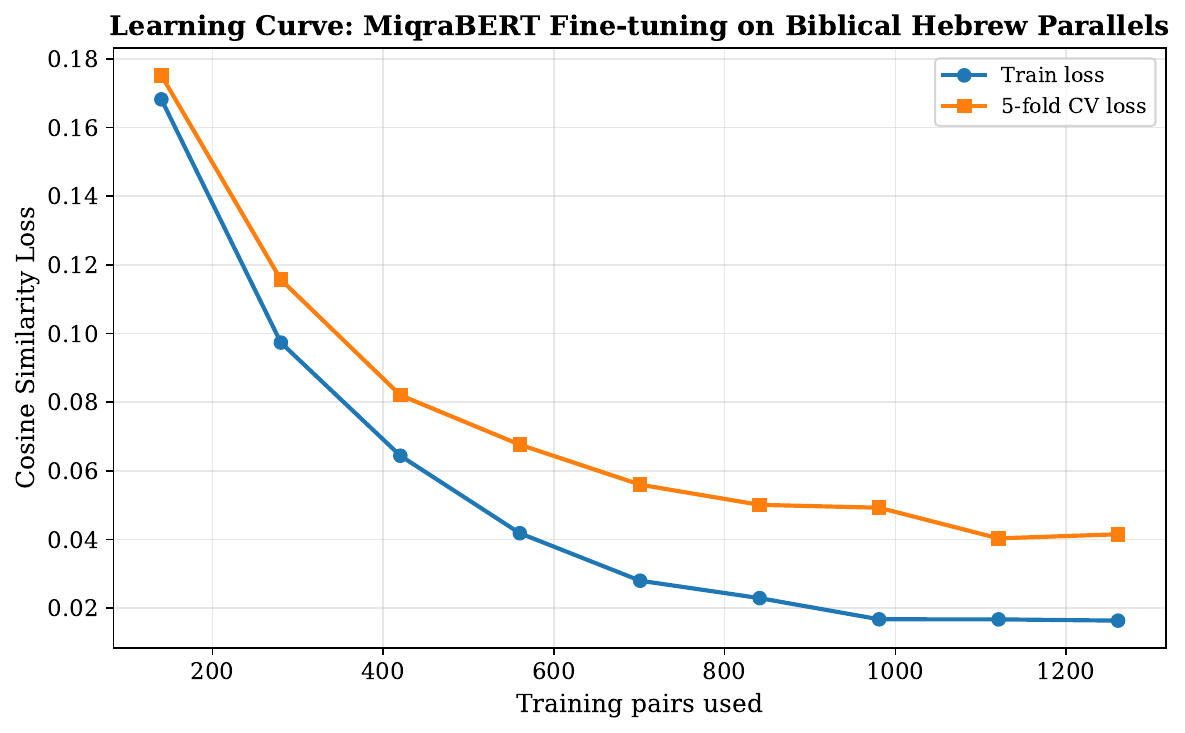}
    \caption{Learning curve for cosine similarity loss during MiqraBERT finetuning. Both training and 5-fold cross-validation loss decrease steeply through approximately 400 pairs, after which the train/CV gap widens.}
    \label{fig:fig4_1_learning_curve_cosloss}
  \end{minipage}\hfill
  \begin{minipage}[t]{0.49\textwidth}
    \centering
    \includegraphics[width=\linewidth]{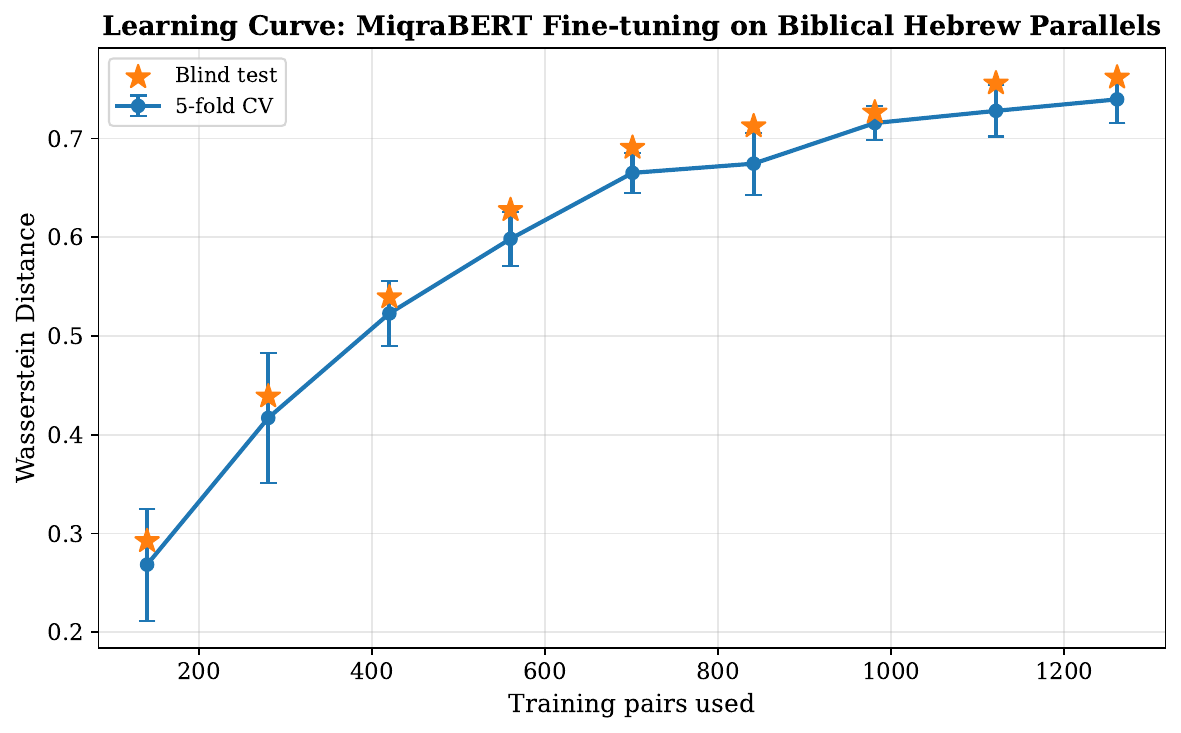}
    \caption{Wasserstein distance as a function of training pairs. Blue circles show 5-fold cross-validation means with error bars; orange stars show blind test set performance. Both metrics plateau above approximately 800 pairs.}
    \label{fig:fig4_2_learning_curve_wsd}
  \end{minipage}
\end{figure}

As shown in Figures 4.1 and 4.2 above, allocations beyond 70\% (1,155 training pairs) yielded diminishing returns. The 90\%-5\%-5\% configuration achieved the highest mean WD (0.776), but this apparent gain is not statistically distinguishable from 70\%-15\%-15\% (\textit{p-value} = 0.086), and the CIs overlap substantially (0.736, 0.766 vs. 0.753, 0.799).\footnote{Statistical comparison via paired t-test across the ten matched seeds, with nine degrees of freedom.} Additionally, the 90\%-5\%-5\% OVL CI (0.027, 0.073) is notably wider than that of 70\%-15\%-15\% (0.049, 0.073), reflecting instability from a test set of just 83 pairs. A similar pattern appears in the 85\%-7.5\%-7.5\% configuration, which achieved lower WD (0.757) than 80\%-10\%-10\% (0.765) despite more training data, alongside the highest OVL in the table (0.077). The diminishing returns are also confirmed in the drop off found around the same split configuration with the cosine similarity loss in Figure 4.1.

The 70\%-15\%-15\% configuration is the optimal choice for this study. It achieves distributional separation statistically indistinguishable from higher training allocations while maintaining a test set of 248 pairs. All subsequent analyses employ the 70\%-15\%-15\% model, and will be known from hereon as MiqraBERT.

\subsection{Embedding Space Transformation}
\label{subsec:embedding-space-transformation}

Before finetuning, AlephBERT compressed both classes into a narrow cosine similarity band spanning significantly around 0.55 to 0.9: this was our anisotropy problem.\footnote{For full metric breakdowns see Smiley, "Intertextual Parallel Detection".} The WD and OVL gains reported in section 4.1 reflect a geometric reorganization of how MiqraBERT now represents BH texts. The finetuned model expanded the distribution of parallel and non-parallel cosine similarity scores nearly threefold. Parallel verses now cluster at higher similarities (mean 0.856 to 0.890 across configurations) while non-parallel verses occupy a distinct lower region (mean 0.114 to 0.137).

\begin{figure}[htbp]
  \centering
  \includegraphics[width=\textwidth]{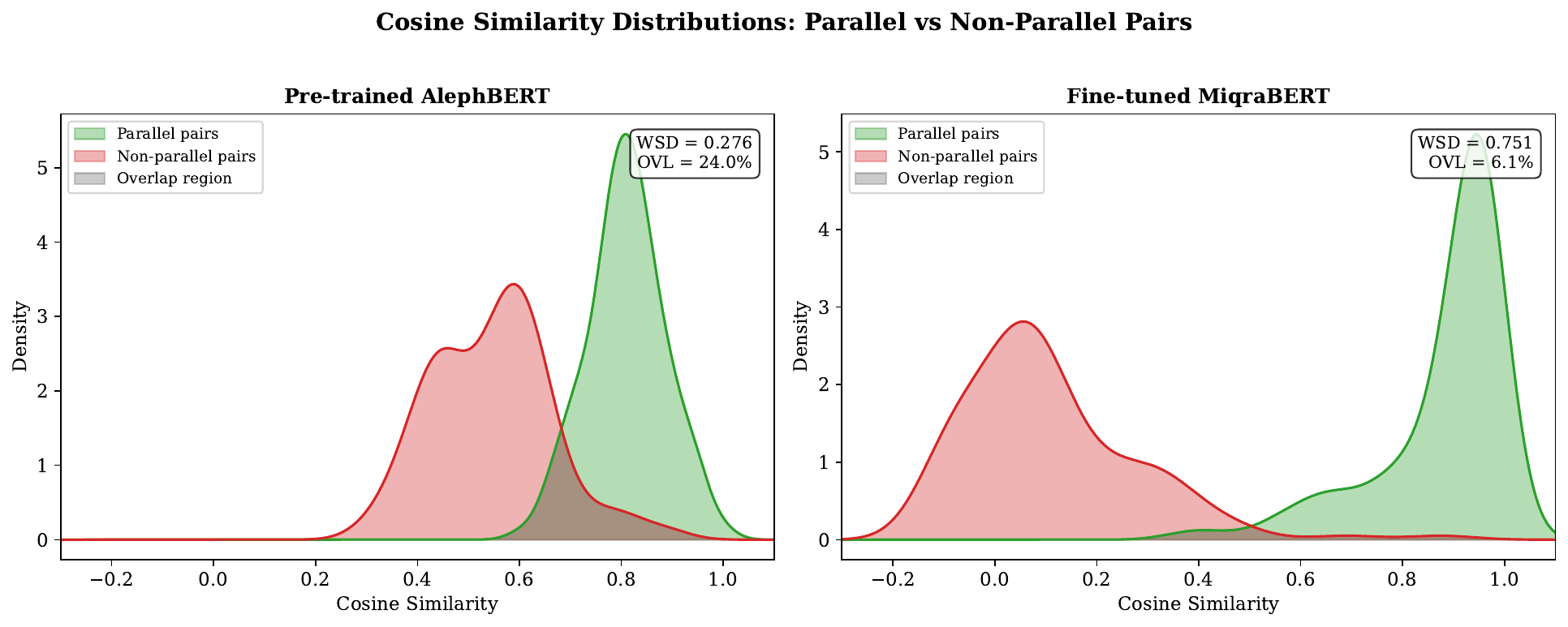}
  \caption{Cosine similarity density distributions for parallel (green) and non-parallel (red) verse pairs before (left) and after (right) finetuning. Gray shading marks the overlap region. Values shown are from a representative single seed; multi-seed averages in the text differ slightly (WD=0.751, OVL=6.1\%).}
  \label{fig:fig4_3_wsd_distribution_shift}
\end{figure}

In the pre-trained embedding space, shown in Figure 4.3, non-parallel scores spread from roughly 0.2 to 0.9 while parallel scores ranged from about 0.5 to 1.0, leaving a wide band where the two classes were indistinguishable (OVL = 24.0\%). After finetuning, the distributions separate into a bimodal pattern with minimal overlap (OVL = 6.1\%). The anisotropic compression that characterized pre-trained AlephBERT, where register distinctions dominated the embedding geometry, has loosened to accommodate for potential semantic content as the space is re-organized around textual similarity with BH examples.

\begin{figure}[htbp]
  \centering
  \includegraphics[width=\textwidth]{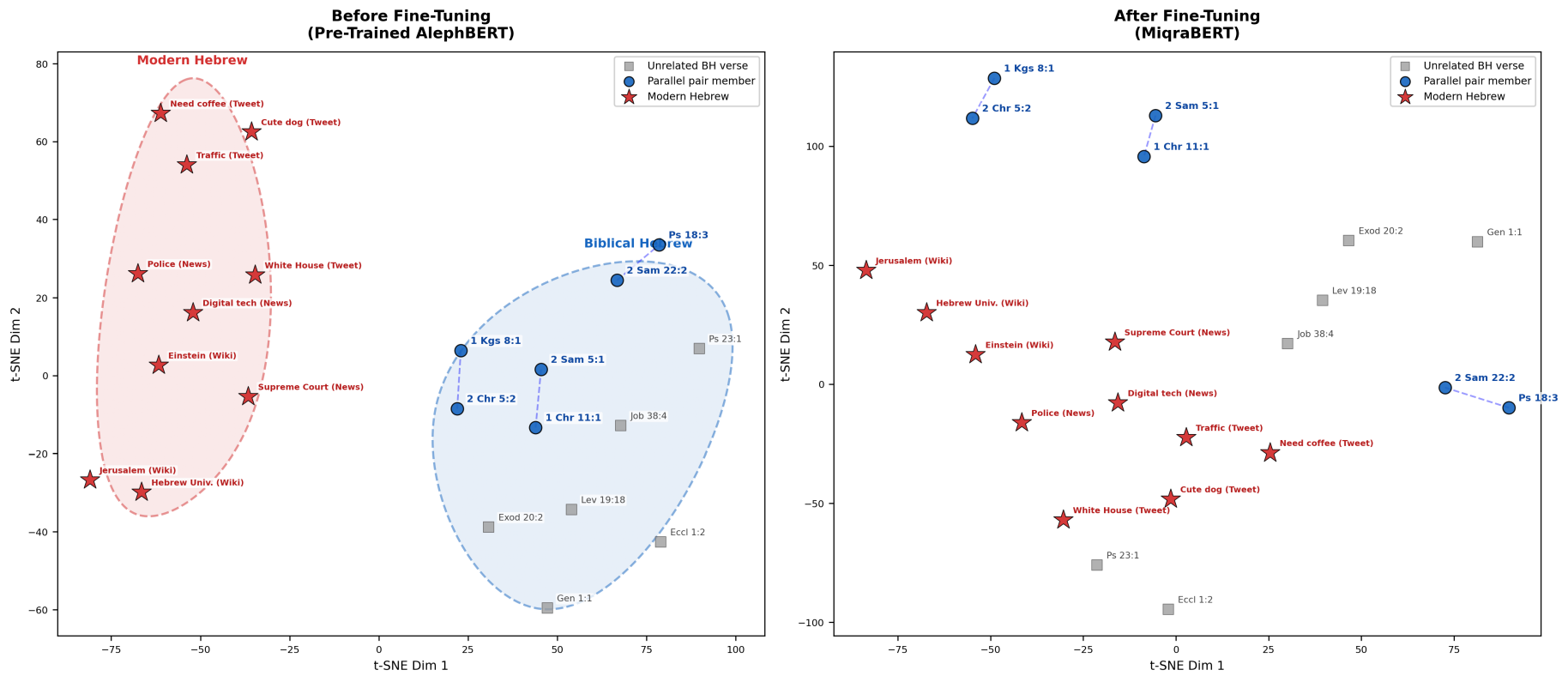}
  \caption{t-SNE projections of sentence embeddings before (left) and after (right) similarity regression finetuning. Blue circles are parallel pair members (connected by dashed lines), gray squares are unrelated BH verses, and red stars are Modern Hebrew texts. After finetuning, parallel pairs cluster together while the rigid BH/MH register boundary dissolves.}
  \label{fig:fig4_4_tsne_before_after}
\end{figure}

Figure 4.4 confirms this decompression at the level of individual verses. In the pre-trained space (left panel), MH texts cluster tightly while BH verses spread across a single undifferentiated region. After finetuning (right panel), parallel pairs such as 2 Sam 22:2 and Ps 18:3 drift together regardless of their original position, and unrelated verses scatter across the embedding space. The register boundary between BH and MH is no longer the dominant organizing axis. By way of our training pairs, MiqraBERT has learned to prioritize textual similarity over the BH/MH linguistic distinctions that dominate the pre-trained model.

\subsection{Qualitative Analysis of Representative Verse Pairs}
\label{subsec:qualitative-analysis-of-representative-verse-pairs}

Table 4.2 presents similarity scores for verse pairs spanning genuine parallels, formulaic overlap, thematic connection, and unrelated texts.\footnote{See Appendix 2 for the Hebrew text of each passage.} From these samples, AlephBERT assigns genuine synoptic parallels scores between 0.87 and 1.00, which looks adequate until one notices that unrelated pairs land between 0.35 and 0.56. The 2 Sam 22:1/Ps 18:1 pair at 0.95 sits only 0.39 above the Song 1:2/Deut 22:5 pair at 0.56, despite the former being a near-verbatim parallel and the latter sharing no relationship whatsoever. MiqraBERT pushes non-parallel scores toward zero (and even in the case of 2 Sam 11:2/Job 38:4 a negative score) while preserving high scores for genuine parallels (0.76 to 1.00), widening this gap from 0.39 to 0.67. The two pairs scoring 1.00 in both columns, 2 Kgs 19:1/Isa 37:1 and Ex 20:2/Deut 5:6, represent identical texts.

\begin{table}[htbp]
  \centering
  \caption{Similarity Scores for Representative Verse Pairs}
  \label{tab:table-42}
  \small
  \begin{tabular}{llp{4.2cm}cc}
    \toprule
    Verse A & Verse B & Relationship Type & AlephBERT & MiqraBERT \\
    \midrule
    2 Sam 22:1 & Ps 18:1 & Synoptic Parallel (Near-Verbatim) & 0.95 & 0.91 \\
    2 Sam 24:1 & 1 Chr 21:1 & Synoptic Parallel (Theological Variation) & 0.87 & 0.76 \\
    2 Kgs 19:1 & Isa 37:1 & Synoptic Parallel (Narrative) & 1.00 & 1.00 \\
    Exod 20:2 & Deut 5:6 & Synoptic Parallel (Decalogue) & 1.00 & 1.00 \\
    Isa 1:2 & Jer 2:4 & Formulaic Overlap (Prophetic Summons) & 0.61 & 0.04 \\
    Ps 23:1 & Ezek 34:2 & Thematic (Shepherd Imagery) & 0.63 & 0.02 \\
    Prov 1:7 & Qoh 12:13 & Thematic (Fear of the LORD) & 0.54 & 0.14 \\
    2 Sam 11:2 & Job 38:4 & No relationship & 0.48 & $-$0.11 \\
    Gen 1:1 & Lev 13:2 & No relationship & 0.35 & 0.06 \\
    Song 1:2 & Deut 22:5 & No relationship & 0.56 & 0.24 \\
    \bottomrule
  \end{tabular}
\end{table}

The 2 Sam 24:1/1 Chr 21:1 pair tests recognition of parallelism despite theological revision, where the Chronicler substitutes "Satan" for YHWH as instigator of David's census.\footnote{See Gary N. Knoppers,\textit{ I Chronicles 10–29: A New Translation with Introduction and Commentary} (AB 12A; New York: Doubleday, 2004), 743–4, 751.} MiqraBERT scores this pair at 0.76, indicating that the model recognizes underlying parallelism through shared narrative structure and semantic overlap despite the lexical substitutions.

Formulaic overlap without a genuine relationship presents a different challenge. Isa 1:2 and Jer 2:4 both open divine speeches with the imperative \texthebrew{שִׁמְעוּ} and calling upon covenant witnesses. AlephBERT scores this pair at 0.61, reflecting shared prophetic conventions. However, these verses introduce entirely different subjects: the heavens and earth for Isaiah and the children of Jacob and clans of Israel in Jeremiah. MiqraBERT reduces these to a score of 0.04 and weighs the lexical differences (e.g. spatial entities vs. people groups) over any other shared "prophetic" framework.

Thematic pairs fare similarly as formula overlaps. Ps 23:1/Ezek 34:2 both employ shepherd imagery; however, YHWH is the Good Shepherd in the psalm while the shepherds of Israel are condemned in Ezekiel. MiqraBERT assigns 0.02 for this pair. Song 1:2/Deut 22:5, while sharing no relationship at all, drop from 0.56 to 0.24. Prov 1:7 and Qoh 12:13 both contain the idea of fearing YHWH, but do not contain any other hints of parallelism in the rest of their verses. AlephBERT scored them at 0.54, and MiqraBERT dropped them to 0.14.

Across all thematic and formulaic pairs their scores consistently stay below 0.25 after finetuning. While, regardless of their pre-trained similarity, the genuine synoptic parallels remain above 0.76. Therefore, it appears that MiqraBERT distinguishes parallels from non-parallels effectively, but does not capture broader thematic and formulaic resonance operating above the level of direct textual correspondence. These distribution and qualitative results establish that MiqraBERT has overcome anisotropy by reshaping the embedding space. The next question is whether that class separation also translates into retrieval performance.

\subsection{Evaluation Protocol}
\label{subsec:evaluation-protocol}

The distribution metrics reported above (WD, OVL) measure how well the model separates parallel from non-parallel pairs within a curated evaluation set. They answer the question: does the finetuned embedding space organize itself so that the two classes occupy distinct regions? Retrieval evaluation poses a different and harder question: given a verse to query with every other verse/half-verse in the HB as potential parallel candidates, can MiqraBERT cast a fine enough net to capture the true parallel? A model could achieve strong distributional separation on balanced test pairs, yet fail at retrieval. This occurs when embedding space contains dense neighborhoods of superficially similar verses that crowd out the true match. Therefore, retrieval metrics test whether distributional quality translates into practical utility at corpus scale.\footnote{On the use of recall@k as a standard metric for evaluating embedding-based retrieval, see Nandan Thakur et al., "BEIR: A Heterogeneous Benchmark for Zero-shot Evaluation of Information Retrieval Models," in \textit{Proceedings of the 2021 Conference on Neural Information Processing Systems: Datasets and Benchmarks Track} (NeurIPS, 2021), 4–5.}

While we have established the 70\%-15\%-15\% configuration as the definitive MiqraBERT model, for the sake of a comprehensive evaluation, as before seven finetuned models spanning training allocations from 50\% to 90\% underwent retrieval evaluation. Each model encoded the complete HB at both verse and half-verse levels, according to the BHSA dataset, and produced a combined embedding matrix of 68,125 vectors, representing all verses and half-verses. Using cosine similarity search each verse/half-verse vector receives a pairwise similarity score with every other verse/half-verse.

Evaluation pairs are derived from the same train-validation-test splits used during finetuning, so that each model faces queries from its own test dataset. For each positive pair (verse A, verse B), the protocol queried with verse A and recorded the rank of verse B among all 68,125 vectors in the corpus. Negative pairs provide a false positive rate estimation, where querying verse A checks whether the unrelated verse B appeared in the top-k results.

\subsection{Retrieval Performance Across Training Allocations}
\label{subsec:retrieval-performance-across-training-allocations}

\begin{table}[htbp]
  \centering
  \caption{Recall@k for All Parallel Pairs by Training Allocation}
  \label{tab:table-43}
  \begin{tabular}{lcccc}
    \toprule
    Training \% & Recall@1 & Recall@5 & Recall@10 & Recall@25 \\
    \midrule
    50 & 0.457 & 0.699 & 0.718 & 0.751 \\
    60 & 0.465 & 0.691 & 0.722 & 0.753 \\
    70 & 0.465 & 0.699 & 0.728 & 0.757 \\
    75 & 0.456 & 0.701 & 0.734 & 0.763 \\
    80 & 0.462 & 0.706 & 0.732 & 0.768 \\
    85 & 0.475 & 0.716 & 0.740 & 0.768 \\
    90 & 0.475 & 0.724 & 0.747 & 0.776 \\
    \bottomrule
  \end{tabular}
\end{table}

Retrieval performance improved modestly with increased training data, although the gains are minimal in absolute terms. The 90\% training allocation achieved the highest recall at every \textit{k} value (tying with 85\% at \textit{k}=1), with recall@10 of 74.7\% and recall@25 of 77.6\%. This is only a 1.9\% difference for both recall@10 and recall@25 for the 70\% model that has been selected as MiqraBERT. Since there are marginal retrieval gains from 70\% to 90\%, these findings do not offset the instability observed in section 4.1, where models above 70\% produce CIs with too wide of a range.

False positive rates remained at or near zero across all configurations. Randomly sampled negative pairs never appeared among the top-25 nearest neighbors for any query, confirming that finetuned models reliably separate genuinely similar verses from random negatives.

\subsection{The Narrative-Poetic Disparity}
\label{subsec:the-narrative-poetic-disparity}

While the embedding and retrieval results are strong enough to warrant MiqraBERT's use, a genre-stratified analysis exposes a performance asymmetry invisible to the aggregate metrics reported above and will constrain the scope of any future study relying on the current version of MiqraBERT for parallel detection. Table 4.4 reports genre-stratified recall@k for the 70\%-15\%-15\% model.

\begin{table}[htbp]
  \centering
  \caption{Recall@k by Parallel Type (70\% Model)}
  \label{tab:table-44}
  \begin{tabular}{lcccc}
    \toprule
    Parallel Genre & Recall@1 & Recall@5 & Recall@10 & Recall@25 \\
    \midrule
    Narrative & 0.558 & 0.842 & 0.871 & 0.905 \\
    Poetic & 0.048 & 0.056 & 0.089 & 0.097 \\
    Difference & +0.510 & +0.786 & +0.782 & +0.808 \\
    \bottomrule
  \end{tabular}
\end{table}

Narrative parallels achieved a recall@10 of 87.1\%, which places the true parallel within the top ten results for nearly nine of every ten Chronicles/Samuel-Kings queries. At \textit{k=}25, recall reached 90.5\%. Poetic parallelism, however, produced starkly different results: recall@10 is only 8.9\%, with a very low 9.7\% at \textit{k}=25. The nearly 80\% gap persisted across all training allocations, indicating a structural limitation of Sentence-BERT rather than a randomly selected training dataset.

The disparity follows from how parallelism operates in each genre. Narrative parallels in Chronicles share substantial lexical overlap with its Samuel-Kings counterpart, despite theological revision and stylistic variation. The Chronicler's substitutions operate within recognizable semantic fields: equivalent names interchange, theological motivations shift, but narrative structure and vocabulary persist (e.g. most categories referenced by Kalimi). Embedding similarity captures these relationships because surface features correlate with genuine parallelism. A TF-IDF baseline bears this out in multiple tests. At the word-level, cosine similarity produces a narrative parallel mean of only 0.675 compared to MiqraBERT's 0.884. Shared vocabulary accounts for much of the signal, but not all of it, and the 0.209 gap represents the structural information that the transformer encodes beyond raw co-occurrence.

Poetic parallelism, on the other hand, operates through different mechanisms. Berlin's multidimensional analysis demonstrates that poetic correspondence spans not just lexical pairing, but also contains grammatical, semantic, and phonological aspects.\footnote{See her taxonomies in Berlin, \textit{The Dynamics of Biblical Parallelism}, 31–126.} Synonymous cola in poetry often do not share any lexemes, yet maintain semantic equivalence. Cosine similarity between surface representations cannot capture relationships grounded in structure, semantics, or sound-play without lexical correspondence.

Meaning, MiqraBERT's verse-level representations, which come from the mean-pooled token embeddings, cannot adequately capture the semantic range from the poetic sets. When two pooled vectors have no overlapping token contributions, there are no weights to pull them together. This is not a problem specific to this model or Sentence-BERT finetuning. It reflects a limitation of embeddings trained on co-occurrence data because they encode lexical relatedness and not semantic similarity. Mrkšić et al. explain, in a process they call counter-fitting, that linguistic constraints need to be injected into the post-processing of embeddings "to improve their usefulness for tasks which involve making semantic similarity judgements."\footnote{Nikola Mrkšić et al., "Counter-fitting Word Vectors to Linguistic Constraints," in \textit{Proceedings of the 2016 Conference of the North American Chapter of the Association for Computational Linguistics: Human Language Technologies }(2016), 146; Counter-fitting needs a rich lexical database. Mrkšić et al. used WordNet, which contains 12,802 antonymy and 31,828 synonymy pairs. Something that is not feasible for BH. George Miller, "WordNet: A Lexical Database for English,"\textit{ Communications of the ACM} 38/11 (1995): 39–41.} Therefore, the poetic limits are the expected ceiling of the current architecture, and it is not evidence that the finetuning process itself, especially for narratives, is to be dismissed.

Training data composition amplifies this limitation. Narrative parallels constitute 67\% of positive examples (556 of 825 pairs), providing substantially more signal for synoptic relationships, but even proportional representation might not resolve the mismatch. The mechanisms generating poetic parallelism differ categorically from those generating narrative parallelism, and a single similarity metric may prove inadequate for both. Nevertheless, the clear usefulness of MiqraBERT for narrative parallel detection warrants the development of computational methods to confirm and explore narrative parallelism in the HB.

\section{Discussion}
\label{sec:discussion}

Three findings emerge from MiqraBERT's evaluations. First, similarity regression finetuning of AlephBERT produces a 2.7-fold improvement in distributional separation with only 1,155 training pairs. This confirms that transfer learning from MH to BH is viable, despite their linguistic, cultural, and temporal differences, and the gain is not reducible simply to lexical overlap, since our TF-IDF benchmark, which only captures shared vocabulary, reaches only 60\% of MiqraBERT's distributional separation. Therefore, the finetuned model has learned representations that operate above the word level. Second, retrieval evaluation exposes a genre-dependent performance asymmetry: narrative synoptic parallels reach a high recall@10 of 87.1\%, while poetic parallels stay well below 9\% at recall@10. Third, both distribution-based metrics (e.g. WD and OVL) reveal meaningful differences between model training-test-validation configurations that recall@k classification metrics obscure. The following demonstrates these results using a set of practical examples, which apply MiqraBERT to narrative parallels that were held out from the training dataset.

In particular, the following stress tests MiqraBERT against two three-way parallel passages to see how well its resistance to anisotropy holds. The first example comes from the narrative retelling of Sennacherib's siege of Jerusalem during Hezekiah's reign, which appears in 2 Kings 18–20, 2 Chronicles 32, and Isaiah 36–38. The second is found in the fall of Jerusalem narratives in 2 Kings 24–25, 2 Chronicles 36, and Jeremiah 52. What makes these cases especially interesting is that the Hebrew text of 2 Kings for both examples is, by lexical distance, closer to its Isaiah-Jeremiah parallels than its 2 Chronicles counterparts.\footnote{Of the thirteen Chronicles–Kings pairs in this set, eight appeared in the training data (2 Chr 32:1//2 Kgs 18:13; 2 Chr 32:9//2 Kgs 18:17; 2 Chr 32:11//2 Kgs 18:22; 2 Chr 32:21//2 Kgs 19:35; 2 Chr 36:11//2 Kgs 24:18; 2 Chr 36:12//2 Kgs 24:19; 2 Chr 36:19//2 Kgs 25:9; 2 Chr 36:20//2 Kgs 25:20) and five did not. All Kings–Isaiah/Jeremiah pairs are entirely out-of-sample. The textual alignments here come from Endres et al., \textit{Chronicles and Its Synoptic Parallels}: 307–17, 339–45. See Appendix 1 for a full list with Hebrew text.}

Across thirteen triplets, MiqraBERT assigned a mean cosine similarity of 0.980 to the 2 Kings–Isaiah/Jeremiah pairs and 0.745 to the Kings–Chronicles pairs. The WD between these two distributions is 0.235 and the OVL is 0.077. This indicates a nearly complete separation between the distribution of pairs. Table 5.1 below reports the individual pairwise scores for all triplets under both AlephBERT and MiqraBERT.

\begin{table}[htbp]
  \centering
  \caption{Pairwise Cosine Similarity Scores for Three-Way Parallel Triplets}
  \label{tab:table-51}
  \small
  \begin{tabular}{lcccc}
    \toprule
    & \multicolumn{2}{c}{Kgs $\leftrightarrow$ Chr} & \multicolumn{2}{c}{Kgs $\leftrightarrow$ Isa/Jer} \\
    \cmidrule(lr){2-3} \cmidrule(lr){4-5}
    Verse Triplet & AlephBERT & MiqraBERT & AlephBERT & MiqraBERT \\
    \midrule
    \multicolumn{5}{l}{\textit{Hezekiah Narrative (2 Kings 18--20 / 2 Chronicles 32 / Isaiah 36--38)}} \\
    \addlinespace
    2 Kgs 18:13 / 2 Chr 32:1 / Isa 36:1   & 0.891 & 0.894 & 0.982 & 0.978 \\
    2 Kgs 18:17 / 2 Chr 32:9 / Isa 36:2   & 0.871 & 0.819 & 0.928 & 0.906 \\
    2 Kgs 18:19 / 2 Chr 32:10 / Isa 36:4  & 0.759 & 0.503 & 0.986 & 0.981 \\
    2 Kgs 18:22 / 2 Chr 32:11 / Isa 36:7  & 0.761 & 0.638 & 0.991 & 0.981 \\
    2 Kgs 18:22 / 2 Chr 32:12 / Isa 36:7  & 0.929 & 0.789 & 0.991 & 0.981 \\
    2 Kgs 19:1 / 2 Chr 32:20 / Isa 37:1   & 0.740 & 0.624 & 1.000 & 1.000 \\
    2 Kgs 19:35 / 2 Chr 32:21 / Isa 37:36 & 0.803 & 0.853 & 0.970 & 0.967 \\
    2 Kgs 20:1 / 2 Chr 32:24 / Isa 38:1   & 0.875 & 0.689 & 0.998 & 0.997 \\
    \midrule
    \multicolumn{5}{l}{\textit{Fall of Jerusalem (2 Kings 24--25 / 2 Chronicles 36 / Jeremiah 52)}} \\
    \addlinespace
    2 Kgs 24:18 / 2 Chr 36:11 / Jer 52:1  & 0.931 & 0.969 & 1.000 & 1.000 \\
    2 Kgs 24:19 / 2 Chr 36:12 / Jer 52:2  & 0.794 & 0.850 & 1.000 & 1.000 \\
    2 Kgs 24:20 / 2 Chr 36:16 / Jer 52:3  & 0.766 & 0.383 & 0.974 & 0.974 \\
    2 Kgs 25:9 / 2 Chr 36:19 / Jer 52:13  & 0.874 & 0.842 & 0.997 & 0.997 \\
    2 Kgs 25:20 / 2 Chr 36:20 / Jer 52:26 & 0.796 & 0.833 & 0.975 & 0.973 \\
    \midrule
    Mean & 0.830 & 0.745 & 0.984 & 0.980 \\
    \bottomrule
  \end{tabular}
\end{table}

The comparisons further illustrate what contrastive learning finetuning was able to accomplish in MiqraBERT. Namely, the anisotropic compression that characterized AlephBERT is apparent in the cosine similarity distribution found in the 2 Kings–2 Chronicles pairs. The mean scores for 2 Kings–2 Chronicles was 0.830 and 0.984 for 2 Kings–Isaiah/Jeremiah: a gap of only 0.154. However, MiqraBERT widened that gap by keeping, relatively, the same mean score for the near-verbatim 2 Kings–Isaiah/Jeremiah parallels and recognizing the differences in the 2 Kings–2 Chronicles passages and lowering most scores when compared to AlephBERT. Note that the pair 2 Kgs 24:20-2 Chr 36:16 went from 0.766 to 0.383. The overall effect is a 52\% improvement (0.154→0.235) in MiqraBERT's ability to separate abbreviated and near-verbatim retellings.

\begin{figure}[htbp]
  \centering
  \includegraphics[width=\textwidth]{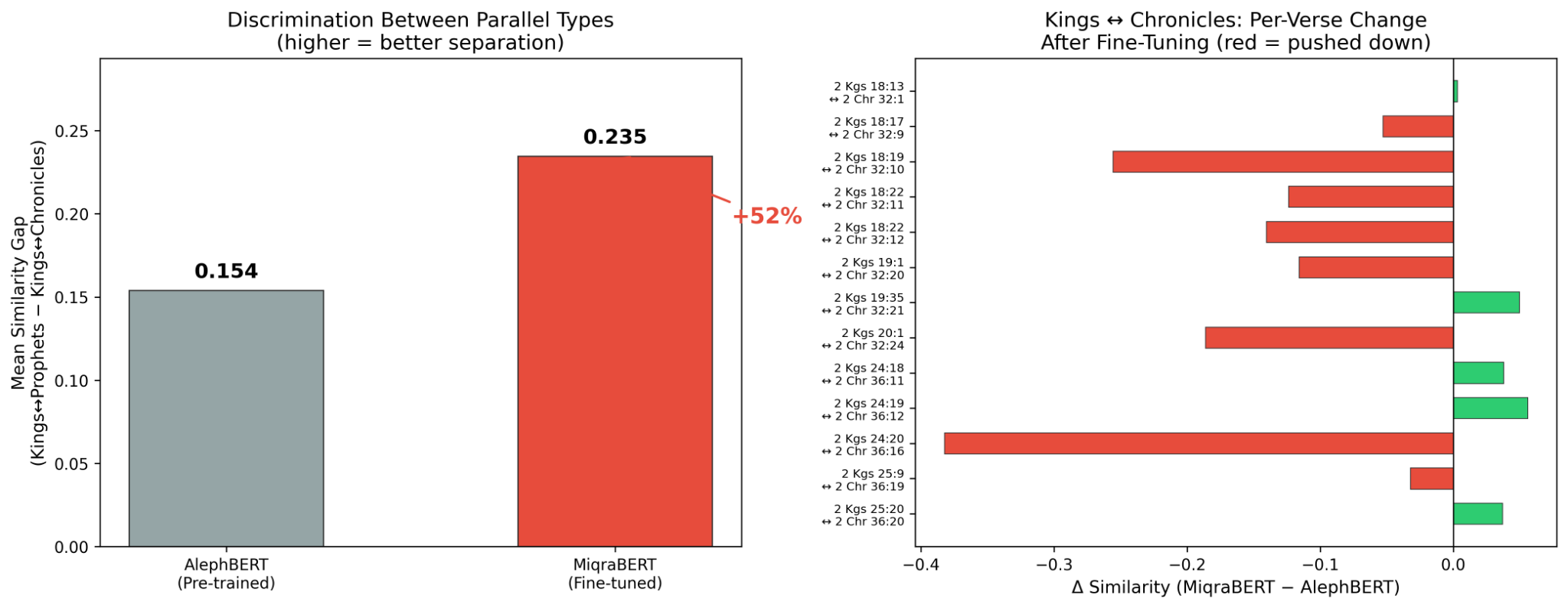}
  \caption{Left: Mean discrimination gap between Kings–Isaiah/Jeremiah and Kings–Chronicles similarity scores for AlephBERT (pre-trained) and MiqraBERT (finetuned). Right: Per-verse change in Kings–Chronicles cosine similarity after finetuning. Red bars indicate decreased scores; green bars indicate modest increases.}
  \label{fig:fig5_1_prophetic_kings_baseline_comparison}
\end{figure}

It is not an accident that 2 Chr 36:16 registers as the most distant pair in the set. 2 Kgs 24:20 and Jer 52:3 share an identical text, except a \textit{mater lectionis }difference in \texthebrew{אוֺתָם}. MiqraBERT scores them at 0.974. The Chronicler, however, has emphasized the mocking of YHWH's messengers, the despising of prophetic words, and the divine rage. The two passages share references to the anger of God, albeit in different phrasing: \texthebrew{אַף יְהוָה} in 2 Kings and Jeremiah and \texthebrew{חֲמַת־יְהוָה} in 2 Chronicles. However, there are no other shared lexemes (see Appendix 1 for the full Hebrew text). MiqraBERT registers this score more accurately at 0.383, while pre-trained AlephBERT scored it at 0.766. Meaning for AlephBERT, it was unable to distinguish the Chronicler's reinterpretation from the closer retellings.

The Rabshakeh episode shows a different mode of the Chronicler's intervention. 2 Kgs 18:19 opens an extended speech to Hezekiah's officials, where Isa 36:4 preserves this address nearly verbatim, and MiqraBERT scores the pair at 0.981. The Chronicler intensifies the taunt by shifting the speaker from an Assyrian official to King Sennacherib himself. The shared lexemes are those of the king of Assyria (\texthebrew{מֶלֶךְ אַשּׁוּר}) and the idea of trust (\texthebrew{בֹּטְחִים/בָּטָחְתָּ}). The switch produced an accurate, low score of 0.503. AlephBERT scored the same pair at 0.759. As with the previous example, this pair was absent from the training data, yet MiqraBERT correctly identified the degree of editorial distance between the source and its retelling.

\begin{figure}[htbp]
  \centering
  \includegraphics[width=\textwidth]{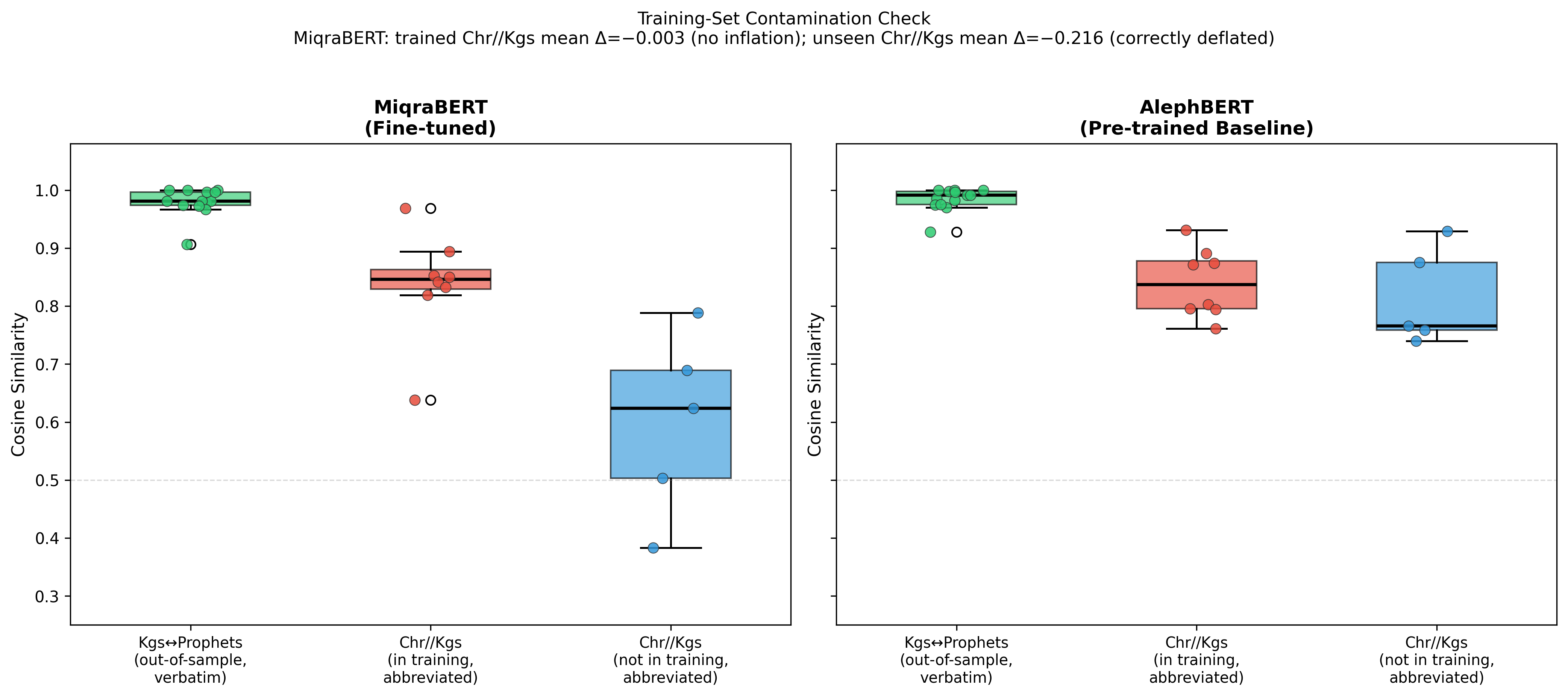}
  \caption{Training-set check. Three-group boxplots for MiqraBERT (left) and AlephBERT (right): Kings–Isaiah/Jeremiah pairs (green, all out-of-sample), Chronicles–Kings pairs seen during training (red), and Chronicles–Kings pairs not in training (blue). MiqraBERT separates all three groups; AlephBERT compresses the two Chronicles–Kings subgroups into an overlapping band.}
  \label{fig:fig5_2_prophetic_kings_contamination_check}
\end{figure}

These examples affirm MiqraBERT's performance reported in sections 4.1–4.4. The finetuned model has acquired the capacity to measure textual similarity from examples given in Chr–Sam/Kgs and transfer that ability to general narrative texts outside of its training dataset. This generalization and the degrees of similarity shown here has direct implications for detecting more subtle types of parallelism. Future research with MiqraBERT will attempt such cases.

\FloatBarrier
\section{Conclusion}
\label{sec:conclusion}

MiqraBERT confirms that transfer learning from MH to BH is viable for verse-level parallel detection, even with a training set of just over a thousand pairs. The finetuned model learns representations that operate above the lexical level, and its narrative retrieval performance is strong enough to warrant application to open questions in biblical scholarship. The distribution-based evaluation framework employed here (e.g. pairing WD and OVL with multi-seed validation) exposed performance instabilities that single-seed classification metrics would have missed. For small-dataset finetuning tasks, these metrics offer a more honest picture of model behavior than threshold-dependent alternatives. The genre-dependent asymmetry with poetic parallelism constrains, but does not invalidate, that application. It instead defines a methodological boundary whose resolution is the subject of ongoing experimental work.

This study has evaluated MiqraBERT on the task it was trained for: distinguishing known parallels from non-parallels in curated test sets. The harder question is what happens beyond that controlled setting. Future studies will test the model's intertextual detection across a wider range of texts and tasks, moving from near-verbatim synoptic parallels into proposed literary relationships where the degree of textual correspondence is contested, partial, or distributed across non-contiguous passages. Whether MiqraBERT can detect those traces reliably, across the full spectrum of relatedness that biblical scholars have proposed, is a question that remains to be answered.

\begin{landscape}
\thispagestyle{empty}
\pagestyle{empty}
\section*{Appendix 1: Hebrew Parallel Texts for Prophetic-Kings Triplets}
\label{sec:appendix1}
\footnotesize
\begin{longtable}{p{2.6cm}>{\raggedleft\arraybackslash}p{6.1cm}>{\raggedleft\arraybackslash}p{6.1cm}>{\raggedleft\arraybackslash}p{6.1cm}}
\toprule
References & 2 Kings & 2 Chronicles & Isaiah/Jeremiah \\
\midrule
\endfirsthead
\toprule
References & 2 Kings & 2 Chronicles & Isaiah/Jeremiah \\
\midrule
\endhead
\bottomrule
\endfoot

2 Kgs 18:13 / 2 Chr 32:1 / Isa 36:1 &
\texthebrew{וּבְאַרְבַּע עֶשְׂרֵה שָׁנָה לַמֶּלֶךְ חִזְקִיָּה עָלָה סַנְחֵרִיב מֶלֶךְ־אַשּׁוּר עַל כָּל־עָרֵי יְהוּדָה הַבְּצֻרוֺת וַיִּתְפְּשֵׂם} &
\texthebrew{אַחֲרֵי הַדְּבָרִים וְהָאֱמֶת הָאֵלֶּה בָּא סַנְחֵרִיב מֶלֶךְ־אַשּׁוּר וַיָּבֹא בִיהוּדָה וַיִּחַן עַל־הֶעָרִים הַבְּצֻרוֺת וַיֹּאמֶר לְבִקְעָם אֵלָיו} &
\texthebrew{וַיְהִי בְּאַרְבַּע עֶשְׂרֵה שָׁנָה לַמֶּלֶךְ חִזְקִיָּהוּ עָלָה סַנְחֵרִיב מֶלֶךְ־אַשּׁוּר עַל כָּל־עָרֵי יְהוּדָה הַבְּצֻרוֺת וַיִּתְפְּשֵׂם} \\
\addlinespace

2 Kgs 18:17 / 2 Chr 32:9 / Isa 36:2 &
\texthebrew{וַיִּשְׁלַח מֶלֶךְ־אַשּׁוּר אֶת־תַּרְתָּן וְאֶת־רַב־סָרִיס וְאֶת־רַב־שָׁקֵה מִן־לָכִישׁ אֶל־הַמֶּלֶךְ חִזְקִיָּהוּ בְּחֵיל כָּבֵד יְרוּשָׁלִָם} &
\texthebrew{אַחַר זֶה שָׁלַח סַנְחֵרִיב מֶלֶךְ־אַשּׁוּר עֲבָדָיו יְרוּשָׁלַיְמָה וְהוּא עַל־לָכִישׁ וְכָל־מֶמְשַׁלְתּוֺ עִמּוֺ עַל־יְחִזְקִיָּהוּ מֶלֶךְ יְהוּדָה וְעַל־כָּל־יְהוּדָה אֲשֶׁר בִּירוּשָׁלִַם לֵאמֹר} &
\texthebrew{וַיִּשְׁלַח מֶלֶךְ־אַשּׁוּר אֶת־רַב־שָׁקֵה מִלָּכִישׁ יְרוּשָׁלְַמָה אֶל־הַמֶּלֶךְ חִזְקִיָּהוּ בְּחֵיל כָּבֵד וַיַּעֲמֹד בִּתְעָלַת הַבְּרֵכָה הָעֶלְיוֺנָה בִּמְסִלַּת שְׂדֵה כוֺבֵס} \\
\addlinespace

2 Kgs 18:19 / 2 Chr 32:10 / Isa 36:4 &
\texthebrew{וַיֹּאמֶר אֲלֵהֶם רַב־שָׁקֵה אִמְרוּ־נָא אֶל־חִזְקִיָּהוּ כֹּה־אָמַר הַמֶּלֶךְ הַגָּדוֺל מֶלֶךְ אַשּׁוּר מָה הַבִּטָּחוֺן הַזֶּה אֲשֶׁר בָּטָחְתָּ} &
\texthebrew{כֹּה אָמַר סַנְחֵרִיב מֶלֶךְ אַשּׁוּר עַל־מָה אַתֶּם בֹּטְחִים וְיֹשְׁבִים בְּמָצוֺר בִּירוּשָׁלִָם} &
\texthebrew{וַיֹּאמֶר אֲלֵיהֶם רַב־שָׁקֵה אִמְרוּ־נָא אֶל־חִזְקִיָּהוּ כֹּה־אָמַר הַמֶּלֶךְ הַגָּדוֺל מֶלֶךְ אַשּׁוּר מָה הַבִּטָּחוֺן הַזֶּה אֲשֶׁר בָּטָחְתָּ} \\
\addlinespace

2 Kgs 18:22 / 2 Chr 32:11 / Isa 36:7 &
\texthebrew{וְכִי־תֹאמְרוּן אֵלַי אֶל־יְהוָה אֱלֹהֵינוּ בָּטָחְנוּ הֲלוֺא־הוּא אֲשֶׁר הֵסִיר חִזְקִיָּהוּ אֶת־בָּמֹתָיו וְאֶת־מִזְבְּחֹתָיו וַיֹּאמֶר לִיהוּדָה וְלִירוּשָׁלִַם לִפְנֵי הַמִּזְבֵּחַ הַזֶּה תִּשְׁתַּחֲווּ בִּירוּשָׁלִָם} &
\texthebrew{הֲלֹא יְחִזְקִיָּהוּ מַסִּית אֶתְכֶם לָתֵת אֶתְכֶם לָמוּת בְּרָעָב וּבְצָמָא לֵאמֹר יְהוָה אֱלֹהֵינוּ יַצִּילֵנוּ מִכַּף מֶלֶךְ אַשּׁוּר} &
\texthebrew{וְכִי־תֹאמַר אֵלַי אֶל־יְהוָה אֱלֹהֵינוּ בָּטָחְנוּ הֲלוֺא־הוּא אֲשֶׁר הֵסִיר חִזְקִיָּהוּ אֶת־בָּמֹתָיו וְאֶת־מִזְבְּחֹתָיו וַיֹּאמֶר לִיהוּדָה וְלִירוּשָׁלִַם לִפְנֵי הַמִּזְבֵּחַ הַזֶּה תִּשְׁתַּחֲווּ} \\
\addlinespace

2 Kgs 18:22 / 2 Chr 32:12 / Isa 36:7 &
\texthebrew{וְכִי־תֹאמְרוּן אֵלַי אֶל־יְהוָה אֱלֹהֵינוּ בָּטָחְנוּ הֲלוֺא־הוּא אֲשֶׁר הֵסִיר חִזְקִיָּהוּ אֶת־בָּמֹתָיו וְאֶת־מִזְבְּחֹתָיו וַיֹּאמֶר לִיהוּדָה וְלִירוּשָׁלִַם לִפְנֵי הַמִּזְבֵּחַ הַזֶּה תִּשְׁתַּחֲווּ בִּירוּשָׁלִָם} &
\texthebrew{הֲלֹא־הוּא יְחִזְקִיָּהוּ הֵסִיר אֶת־בָּמֹתָיו וְאֶת־מִזְבְּחֹתָיו וַיֹּאמֶר לִיהוּדָה וְלִירוּשָׁלִַם לֵאמֹר לִפְנֵי מִזְבֵּחַ אֶחָד תִּשְׁתַּחֲווּ וְעָלָיו תַּקְטִירוּ} &
\texthebrew{וְכִי־תֹאמַר אֵלַי אֶל־יְהוָה אֱלֹהֵינוּ בָּטָחְנוּ הֲלוֺא־הוּא אֲשֶׁר הֵסִיר חִזְקִיָּהוּ אֶת־בָּמֹתָיו וְאֶת־מִזְבְּחֹתָיו וַיֹּאמֶר לִיהוּדָה וְלִירוּשָׁלִַם לִפְנֵי הַמִּזְבֵּחַ הַזֶּה תִּשְׁתַּחֲווּ} \\
\addlinespace

2 Kgs 19:1 / 2 Chr 32:20 / Isa 37:1 &
\texthebrew{וַיְהִי כִּשְׁמֹעַ הַמֶּלֶךְ חִזְקִיָּהוּ וַיִּקְרַע אֶת־בְּגָדָיו וַיִּתְכַּס בַּשָּׂק וַיָּבֹא בֵּית יְהוָה} &
\texthebrew{וַיִּתְפַּלֵּל יְחִזְקִיָּהוּ הַמֶּלֶךְ וִישַׁעְיָהוּ בֶן־אָמוֺץ הַנָּבִיא עַל־זֹאת וַיִּזְעֲקוּ הַשָּׁמָיִם} &
\texthebrew{וַיְהִי כִּשְׁמֹעַ הַמֶּלֶךְ חִזְקִיָּהוּ וַיִּקְרַע אֶת־בְּגָדָיו וַיִּתְכַּס בַּשָּׂק וַיָּבֹא בֵּית יְהוָה} \\
\addlinespace

2 Kgs 19:35 / 2 Chr 32:21 / Isa 37:36 &
\texthebrew{וַיְהִי בַּלַּיְלָה הַהוּא וַיֵּצֵא מַלְאַךְ יְהוָה וַיַּךְ בְּמַחֲנֵה אַשּׁוּר מֵאָה שְׁמוֺנִים וַחֲמִשָּׁה אָלֶף וַיַּשְׁכִּימוּ בַבֹּקֶר וְהִנֵּה כֻלָּם פְּגָרִים מֵתִים} &
\texthebrew{וַיִּשְׁלַח יְהוָה מַלְאָךְ וַיַּכְחֵד כָּל־גִּבּוֺר חַיִל וְנָגִיד וְשָׂר בְּמַחֲנֵה מֶלֶךְ אַשּׁוּר וַיָּשָׁב בְּבֹשֶׁת פָּנִים לְאַרְצוֺ וַיָּבֹא בֵּית אֱלֹהָיו וּמִיצִיאֵי מֵעָיו שָׁם הִפִּילֻהוּ בֶחָרֶב} &
\texthebrew{וַיֵּצֵא מַלְאַךְ יְהוָה וַיַּכֶּה בְּמַחֲנֵה אַשּׁוּר מֵאָה וּשְׁמֹנִים וַחֲמִשָּׁה אָלֶף וַיַּשְׁכִּימוּ בַבֹּקֶר וְהִנֵּה כֻלָּם פְּגָרִים מֵתִים} \\
\addlinespace

2 Kgs 20:1 / 2 Chr 32:24 / Isa 38:1 &
\texthebrew{בַּיָּמִים הָהֵם חָלָה חִזְקִיָּהוּ לָמוּת וַיָּבֹא אֵלָיו יְשַׁעְיָהוּ בֶן־אָמוֺץ הַנָּבִיא וַיֹּאמֶר אֵלָיו כֹּה־אָמַר יְהוָה צַו לְבֵיתֶךָ כִּי מֵת אַתָּה וְלֹא תִחְיֶה} &
\texthebrew{בַּיָּמִים הָהֵם חָלָה יְחִזְקִיָּהוּ עַד־לָמוּת וַיִּתְפַּלֵּל אֶל־יְהוָה וַיֹּאמֶר לוֺ וּמוֺפֵת נָתַן לוֺ} &
\texthebrew{בַּיָּמִים הָהֵם חָלָה חִזְקִיָּהוּ לָמוּת וַיָּבוֺא אֵלָיו יְשַׁעְיָהוּ בֶן־אָמוֺץ הַנָּבִיא וַיֹּאמֶר אֵלָיו כֹּה־אָמַר יְהוָה צַו לְבֵיתֶךָ כִּי מֵת אַתָּה וְלֹא תִחְיֶה} \\
\midrule

2 Kgs 24:18 / 2 Chr 36:11 / Jer 52:1 &
\texthebrew{בֶּן־עֶשְׂרִים וְאַחַת שָׁנָה צִדְקִיָּהוּ בְמָלְכוֺ וְאַחַת עֶשְׂרֵה שָׁנָה מָלַךְ בִּירוּשָׁלִָם וְשֵׁם אִמּוֺ חֲמוּטַל בַּת־יִרְמְיָהוּ מִלִּבְנָה} &
\texthebrew{בֶּן־עֶשְׂרִים וְאַחַת שָׁנָה צִדְקִיָּהוּ בְמָלְכוֺ וְאַחַת עֶשְׂרֵה שָׁנָה מָלַךְ בִּירוּשָׁלִָם} &
\texthebrew{בֶּן־עֶשְׂרִים וְאַחַת שָׁנָה צִדְקִיָּהוּ בְמָלְכוֺ וְאַחַת עֶשְׂרֵה שָׁנָה מָלַךְ בִּירוּשָׁלִָם וְשֵׁם אִמּוֺ חֲמוּטַל בַּת־יִרְמְיָהוּ מִלִּבְנָה} \\
\addlinespace

2 Kgs 24:19 / 2 Chr 36:12 / Jer 52:2 &
\texthebrew{וַיַּעַשׂ הָרַע בְּעֵינֵי יְהוָה כְּכֹל אֲשֶׁר־עָשָׂה יְהוֺיָקִים} &
\texthebrew{וַיַּעַשׂ הָרַע בְּעֵינֵי יְהוָה אֱלֹהָיו לֹא נִכְנַע מִלִּפְנֵי יִרְמְיָהוּ הַנָּבִיא מִפִּי יְהוָה} &
\texthebrew{וַיַּעַשׂ הָרַע בְּעֵינֵי יְהוָה כְּכֹל אֲשֶׁר־עָשָׂה יְהוֺיָקִים} \\
\addlinespace

2 Kgs 24:20 / 2 Chr 36:16 / Jer 52:3 &
\texthebrew{כִּי עַל־אַף יְהוָה הָיְתָה בִירוּשָׁלִַם וּבִיהוּדָה עַד־הִשְׁלִכוֺ אֹתָם מֵעַל פָּנָיו וַיִּמְרֹד צִדְקִיָּהוּ בְּמֶלֶךְ בָּבֶל} &
\texthebrew{וַיִּהְיוּ מַלְעִבִים בְּמַלְאֲכֵי הָאֱלֹהִים וּבוֺזִים דְּבָרָיו וּמִתַּעְתְּעִים בִּנְבִאָיו עַד עֲלוֺת חֲמַת־יְהוָה בְּעַמּוֺ עַד־לְאֵין מַרְפֵּא} &
\texthebrew{כִּי עַל־אַף יְהוָה הָיְתָה בִּירוּשָׁלִַם וִיהוּדָה עַד־הִשְׁלִיכוֺ אוֺתָם מֵעַל פָּנָיו וַיִּמְרֹד צִדְקִיָּהוּ בְּמֶלֶךְ בָּבֶל} \\
\addlinespace

2 Kgs 25:9 / 2 Chr 36:19 / Jer 52:13 &
\texthebrew{וַיִּשְׂרֹף אֶת־בֵּית־יְהוָה וְאֶת־בֵּית הַמֶּלֶךְ וְאֵת כָּל־בָּתֵּי יְרוּשָׁלִַם וְאֶת־כָּל־בֵּית גָּדוֺל שָׂרַף בָּאֵשׁ} &
\texthebrew{וַיִּשְׂרְפוּ אֶת־בֵּית הָאֱלֹהִים וַיְנַתְּצוּ אֵת חוֺמַת יְרוּשָׁלִָם וְכָל־אַרְמְנוֺתֶיהָ שָׂרְפוּ בָאֵשׁ וְכָל־כְּלֵי מַחֲמַדֶּיהָ לְהַשְׁחִית} &
\texthebrew{וַיִּשְׂרֹף אֶת־בֵּית־יְהוָה וְאֶת־בֵּית הַמֶּלֶךְ וְאֵת כָּל־בָּתֵּי יְרוּשָׁלִַם וְאֶת־כָּל־בֵּית הַגָּדוֺל שָׂרַף בָּאֵשׁ} \\
\addlinespace

2 Kgs 25:20 / 2 Chr 36:20 / Jer 52:26 &
\texthebrew{וַיִּקַּח אֹתָם נְבוּזַרְאֲדָן רַב־טַבָּחִים וַיֹּלֶךְ אֹתָם עַל־מֶלֶךְ בָּבֶל רִבְלָתָה} &
\texthebrew{וַיֶּגֶל הַשְּׁאֵרִית מִן־הַחֶרֶב אֶל־בָּבֶל וַיִּהְיוּ־לוֺ וּלְבָנָיו לַעֲבָדִים עַד־מְלֹךְ מַלְכוּת פָּרָס} &
\texthebrew{וַיִּקַּח אוֺתָם נְבוּזַרְאֲדָן רַב־טַבָּחִים וַיֹּלֶךְ אוֺתָם אֶל־מֶלֶךְ בָּבֶל רִבְלָתָה} \\

\end{longtable}
\end{landscape}

\begin{landscape}
\thispagestyle{empty}
\pagestyle{empty}
\section*{Appendix 2: Hebrew Texts for the Representative Verse Pairs}
\label{sec:appendix2}
\footnotesize
\begin{longtable}{p{3cm}>{\raggedleft\arraybackslash}p{8.6cm}>{\raggedleft\arraybackslash}p{8.6cm}}
\toprule
References A / B & Verse A & Verse B \\
\midrule
\endfirsthead
\toprule
References A / B & Verse A & Verse B \\
\midrule
\endhead
\bottomrule
\endfoot

2 Sam 22:1 / Ps 18:1 &
\texthebrew{וַיְדַבֵּר דָּוִד לַיהוָה אֶת־דִּבְרֵי הַשִּׁירָה הַזֹּאת בְּיוֹם הִצִּיל יְהוָה אֹתוֹ מִכַּף כָּל־אֹיְבָיו וּמִכַּף שָׁאוּל} &
\texthebrew{לַמְנַצֵּחַ לְעֶבֶד יְהוָה לְדָוִד אֲשֶׁר דִּבֶּר לַיהוָה אֶת־דִּבְרֵי הַשִּׁירָה הַזֹּאת בְּיוֹם הִצִּיל־יְהוָה אוֹתוֹ מִכַּף כָּל־אֹיְבָיו וּמִיַּד שָׁאוּל} \\
\addlinespace

2 Sam 24:1 / 1 Chr 21:1 &
\texthebrew{וַיֹּסֶף אַף־יְהוָה לַחֲרוֹת בְּיִשְׂרָאֵל וַיָּסֶת אֶת־דָּוִד בָּהֶם לֵאמֹר לֵךְ מְנֵה אֶת־יִשְׂרָאֵל וְאֶת־יְהוּדָה} &
\texthebrew{וַיַּעֲמֹד שָׂטָן עַל־יִשְׂרָאֵל וַיָּסֶת אֶת־דָּוִיד לִמְנוֹת אֶת־יִשְׂרָאֵל} \\
\addlinespace

2 Kgs 19:1 / Isa 37:1 &
\texthebrew{וַיְהִי כִּשְׁמֹעַ הַמֶּלֶךְ חִזְקִיָּהוּ וַיִּקְרַע אֶת־בְּגָדָיו וַיִּתְכַּס בַּשָּׂק וַיָּבֹא בֵּית יְהוָה} &
\texthebrew{וַיְהִי כִּשְׁמֹעַ הַמֶּלֶךְ חִזְקִיָּהוּ וַיִּקְרַע אֶת־בְּגָדָיו וַיִּתְכַּס בַּשָּׂק וַיָּבֹא בֵּית יְהוָה} \\
\addlinespace

Exod 20:2 / Deut 5:6 &
\texthebrew{אָנֹכִי יְהוָה אֱלֹהֶיךָ אֲשֶׁר הוֹצֵאתִיךָ מֵאֶרֶץ מִצְרַיִם מִבֵּית עֲבָדִים} &
\texthebrew{אָנֹכִי יְהוָה אֱלֹהֶיךָ אֲשֶׁר הוֹצֵאתִיךָ מֵאֶרֶץ מִצְרַיִם מִבֵּית עֲבָדִים} \\
\addlinespace

Isa 1:2 / Jer 2:4 &
\texthebrew{שִׁמְעוּ שָׁמַיִם וְהַאֲזִינִי אֶרֶץ כִּי יְהוָה דִּבֵּר בָּנִים גִּדַּלְתִּי וְרוֹמַמְתִּי וְהֵם פָּשְׁעוּ בִי} &
\texthebrew{שִׁמְעוּ דְבַר־יְהוָה בֵּית יַעֲקֹב וְכָל־מִשְׁפְּחוֹת בֵּית יִשְׂרָאֵל} \\
\addlinespace

Ps 23:1 / Ezek 34:2 &
\texthebrew{מִזְמוֹר לְדָוִד יְהוָה רֹעִי לֹא אֶחְסָר} &
\texthebrew{בֶּן־אָדָם הִנָּבֵא עַל־רוֹעֵי יִשְׂרָאֵל הִנָּבֵא וְאָמַרְתָּ אֲלֵיהֶם לָרֹעִים כֹּה אָמַר אֲדֹנָי יְהוִה הוֹי רֹעֵי־יִשְׂרָאֵל אֲשֶׁר הָיוּ רֹעִים אוֹתָם הֲלוֹא הַצֹּאן יִרְעוּ הָרֹעִים} \\
\addlinespace

Prov 1:7 / Qoh 12:13 &
\texthebrew{יִרְאַת יְהוָה רֵאשִׁית דָּעַת חָכְמָה וּמוּסָר אֱוִילִים בָּזוּ} &
\texthebrew{סוֹף דָּבָר הַכֹּל נִשְׁמָע אֶת־הָאֱלֹהִים יְרָא וְאֶת־מִצְוֹתָיו שְׁמוֹר כִּי־זֶה כָּל־הָאָדָם} \\
\addlinespace

2 Sam 11:2 / Job 38:4 &
\texthebrew{וַיְהִי לְעֵת הָעֶרֶב וַיָּקָם דָּוִד מֵעַל מִשְׁכָּבוֹ וַיִּתְהַלֵּךְ עַל־גַּג בֵּית־הַמֶּלֶךְ וַיַּרְא אִשָּׁה רֹחֶצֶת מֵעַל הַגָּג וְהָאִשָּׁה טוֹבַת מַרְאֶה מְאֹד} &
\texthebrew{אֵיפֹה הָיִיתָ בְּיָסְדִי־אָרֶץ הַגֵּד אִם־יָדַעְתָּ בִינָה} \\
\addlinespace

Gen 1:1 / Lev 13:2 &
\texthebrew{בְּרֵאשִׁית בָּרָא אֱלֹהִים אֵת הַשָּׁמַיִם וְאֵת הָאָרֶץ} &
\texthebrew{אָדָם כִּי־יִהְיֶה בְעוֹר־בְּשָׂרוֹ שְׂאֵת אוֹ־סַפַּחַת אוֹ בַהֶרֶת וְהָיָה בְעוֹר־בְּשָׂרוֹ לְנֶגַע צָרָעַת וְהוּבָא אֶל־אַהֲרֹן הַכֹּהֵן אוֹ אֶל־אַחַד מִבָּנָיו הַכֹּהֲנִים} \\
\addlinespace

Song 1:2 / Deut 22:5 &
\texthebrew{יִשָּׁקֵנִי מִנְּשִׁיקוֹת פִּיהוּ כִּי־טוֹבִים דֹּדֶיךָ מִיָּיִן׃} &
\texthebrew{לֹא־יִהְיֶה כְלִי־גֶבֶר עַל־אִשָּׁה וְלֹא־יִלְבַּשׁ גֶּבֶר שִׂמְלַת אִשָּׁה כִּי תוֹעֲבַת יְהוָה אֱלֹהֶיךָ כָּל־עֹשֵׂה אֵלֶּה} \\
\addlinespace

\end{longtable}
\end{landscape}
\pagestyle{plain}

\section*{Bibliography}
\label{sec:bibliography}
\begin{hangparas}{1.5em}{1}

Alter, Robert. \textit{The Art of Biblical Poetry}. 2nd ed. New York: Basic Books, 2011.

Berlin, Adele. \textit{The Dynamics of Biblical Parallelism}. Revised and Expanded Edition. Grand Rapids: Eerdmans, 2008.

Biś, Daniel, Maksim Podkorytov, and Xiuwen Liu. "Too Much in Common: Shifting of Embeddings in Transformer Language Models and its Implications." Pages 5117–30 in \textit{Proceedings of the 2021 Conference of the North American Chapter of the Association for Computational Linguistics: Human Language Technologies}. Online: Association for Computational Linguistics, 2021.

Bothwell, Stephen, Justin DeBenedetto, Theresa Crnkovich, Hildegund Muller, and David Chiang. "Introducing Rhetorical Parallelism Detection: A New Task with Datasets, Metrics, and Baselines." Pages 5007–39 in \textit{Proceedings of the 2023 Conference on Empirical Methods in Natural Language Processing}. Singapore: Association for Computational Linguistics, 2023.

Brill, Oran, Moshe Koppel, and Avi Shmidman. "FAST: Fast and Accurate Synoptic Texts." \textit{Digital Scholarship in the Humanities} 35/2 (2020): 254–264.

Burns, Patrick J., James A. Brofos, Kyle Li, Pramit Chaudhuri, and Joseph P. Dexter. "Profiling of Intertextuality in Latin Literature Using Word Embeddings." Pages 4900–4907 in \textit{Proceedings of the 2021 Conference of the North American Chapter of the Association for Computational Linguistics: Human Language Technologies}. Association for Computational Linguistics, 2021.

Dodge, Jesse, Gabriel Ilharco, Roy Schwartz, Ali Farhadi, Hannaneh Hajishirzi, and Noah Smith. "Fine-Tuning Pretrained Language Models: Weight Initializations, Data Orders, and Early Stopping." \textit{arXiv:2002.06305}, 2020. \url{https://doi.org/10.48550/arXiv.2002.06305}

Doron, Edit. "The Biblical Sources of Modern Hebrew Syntax." Pages 222–56 in \textit{Language Contact, Continuity and Change in the Genesis of Modern Hebrew}. Amsterdam: John Benjamins Publishing Company, 2019.

Efron, Bradley, and Robert J. Tibshirani. \textit{An Introduction to the Bootstrap}. Monographs on Statistics and Applied Probability 57. New York: Chapman \& Hall, 1993.

Endres, John Carol, Corrine L. Patton, William R. Millar, John Barclay Burns, Corrine L. Carvalho, Pauline A. Viviano, and Jim Fitzgerald, eds. \textit{Chronicles and Its Synoptic Parallels in Samuel, Kings, and Related Biblical Texts}. Collegeville, MN: Liturgical Press, 1998.

Ethayarajh, Kawin. "How Contextual are Contextualized Word Representations? Comparing the Geometry of BERT, ELMo, and GPT-2 Embeddings." Pages 55–65 in \textit{Proceedings of the 2019 Conference on Empirical Methods in Natural Language Processing and the 9th International Joint Conference on Natural Language Processing (EMNLP-IJCNLP)}. Hong Kong, China: Association for Computational Linguistics, 2019.

Evans, Robert. "Shitposting, Inspirational Terrorism, and the Christchurch Mosque Massacre." \textit{Bellingcat}, March 15, 2019. Accessed December 8, 2025. \url{https://www.bellingcat.com/news/rest-of-world/2019/03/15/shitposting-inspirational-terrorism-and-the-christchurch-mosque-massacre/}.

Even-Shoshan, Avraham. \textit{The New Dictionary }[in Hebrew]. Jerusalem: Kirjat Sepher, 1970.

Fawcett, Tom. "An Introduction to ROC Analysis." \textit{Pattern Recognition Letters} 27/8 (2006): 861–74. \url{https://doi.org/10.1016/j.patrec.2005.10.010}

Fokkelman, Joannes Petrus. \textit{Reading Biblical Poetry: An Introductory Guide.} Translated by Ineke Smit. Louisville, KY: Westminster John Knox Press, 2001.

Fono, Niv, Harel Moshayof, Eldar Karol, Itai Assraf, and Mark Last. "Embible: Reconstruction of Ancient Hebrew and Aramaic Texts Using Transformers." Pages 846–52 in \textit{Findings of the Association for Computational Linguistics: EACL 2024}. St. Julian's, Malta: Association for Computational Linguistics, 2024.

Gao, Jun, Di He, Xu Tan, Tao Qin, Liwei Wang, and Tieyan Liu. "Representation Degeneration Problem in Training Natural Language Generation Models." Pages 1–14 in \textit{International Conference on Learning Representations}. New Orleans, LA: OpenReview, 2019.

Gao, Tianyu, Xingcheng Yao, and Danqi Chen. "SimCSE: Simple Contrastive Learning of Sentence Embeddings." Pages 6894–6910 in \textit{Proceedings of the 2021 Conference on Empirical Methods in Natural Language Processing}. Punta Cana, Dominican Republic: Association for Computational Linguistics, 2021.

Godey, Nathan, Éric de la Clergerie, and Benoît Sagot. "Anisotropy Is Inherent to Self-Attention in Transformers." Pages 35–48 in \textit{Proceedings of the 18th Conference of the European Chapter of the Association for Computational Linguistics, Volume 1: Long Papers}. St. Julian's, Malta: Association for Computational Linguistics, 2024.

Gosset, William S. [Student]. "The Probable Error of a Mean." \textit{Biometrika} 6 (1908): 1–25.

HaCohen-Kerner, Yaakov, Hananya Beck, Elchai Yehudai, and Dror Mughaz. "Identifying Historical Period and Ethnic Origin of Documents Using Stylistic Feature Sets." Pages 102–13 in \textit{Discovery Science}. Springer Berlin Heidelberg, 2006.

HaCohen-Kerner, Yaakov, Hananya Beck, Elchai Yehudai, and Dror Mughaz. "Stylistic Feature Sets As Classifiers of Documents According to Their Historical Period and Ethnic Origin." \textit{Applied Artificial Intelligence} 24/9 (2010): 847–62.

HaCohen-Kerner, Yaakov, Nadav Schweitzer, and Dror Mughaz. "Automatically Identifying Citations in Hebrew-Aramaic Documents." \textit{Cybernetics and Systems} 42/3 (2011): 180–97.

Jurafsky, Daniel, and James H. Martin. \textit{Speech and Language Processing: An Introduction to Natural Language Processing, Computational Linguistics, and Speech Recognition, with Language Models}. 3rd ed. Online manuscript released January 6, 2026.

Kalimi, Isaac. \textit{The Reshaping of Ancient Israelite History in Chronicles.} University Park, PA: Penn State University Press, 2012.

Kaplan, Jared, Sam McCandlish, Tom Henighan, Tom B. Brown, Benjamin Chess, Rewon Child, Scott Gray, Alec Radford, Jeffrey Wu, and Dario Amodei. "Scaling Laws for Neural Language Models." \textit{arXiv:2001.08361}, 2020. \url{https://doi.org/10.48550/arXiv.2001.08361}

Knoppers, Gary N. \textit{I Chronicles 10–29: A New Translation with Introduction and Commentary.} Anchor Bible 12A. New York: Doubleday, 2004.

Kugel, James L. \textit{The Idea of Biblical Poetry: Parallelism and Its History}. New Haven: Yale University Press, 1981.

Levenshtein, Vladimir I. "Binary Codes Capable of Correcting Deletions, Insertions, and Reversals." \textit{Soviet Physics Doklady} 10 (1965): 707–710.

Li, Bohan, Hao Zhou, Junxian He, Mingxuan Wang, Yiming Yang, and Lei Li. "On the Sentence Embeddings from Pre-trained Language Models." Pages 9119–30 in\textit{ Proceedings of the 2020 Conference on Empirical Methods in Natural Language Processing (EMNLP)}. Online: Association for Computational Linguistics, 2020.

Li, Qian, Hao Peng, Jianxin Li, Congying Xia, Renyu Yang, Lichao Sun, Philip S. Yu, and Lifang He. "A Survey on Text Classification: From Traditional to Deep Learning." \textit{ACM Transactions on Intelligent Systems and Technology }13/2 (2022): 1–41.

Liebeskind, Chaya, and Shmuel Liebeskind. "Deep Learning for Period Classification of Historical Hebrew Texts."\textit{ Journal of Data Mining \& Digital Humanities} (2020): 5864.

Liebeskind, Chaya, Shmuel Liebeskind, and Dan Bouhnik. "Machine Translation for Historical Research: A Case Study of Aramaic-Ancient Hebrew Translations." \textit{Journal on Computing and Digital Heritage} 17/2 (2024): 1–23.

Lobbezoo, Bert. "Computer-Based Recognition of Intertextuality within the Hebrew Bible." Master's Thesis. Delft University of Technology, 2015.

McGovern, Hope, Hale Sirin, Tom Lippincott, and Andrew Caines. "Detecting Narrative Patterns in Biblical Hebrew and Greek." Pages 269–79 in \textit{Proceedings of the 1st Workshop on Machine Learning for Ancient Languages (ML4AL 2024)}. Association for Computational Linguistics, 2024.

Miller, George A. "WordNet: A Lexical Database for English."\textit{ Communications of the ACM} 38/11 (1995): 39–41.

Miller, Hadar, Tsvi Kuflik, and Moshe Lavee. "Text Alignment in the Service of Text Reuse Detection." \textit{Applied Sciences} 15/6 (2025): 3395.

Mosbach, Marius, Maksym Andriushchenko, and Dietrich Klakow. "On the Stability of Fine-Tuning BERT: Misconceptions, Explanations, and Strong Baselines." \textit{9th International Conference on Learning Representations (ICLR) 2021}. Open Review, 2021.

Mrkšić, Nikola, Diarmuid Ó Séaghdha, Blaise Thomson, Milica Gašić, Lina Rojas-Barahona, Pei-Hao Su, David Vandyke, Tsung-Hsien Wen, and Steve Young. "Counter-fitting Word Vectors to Linguistic Constraints." Pages 142–48in\textit{ Proceedings of the 2016 Conference of the North American Chapter of the Association for Computational Linguistics: Human Language Technologies, San Diego, CA}: Association for Computational Linguistics, 2016.

Mughaz, Dror, Yaakov HaCohen-Kerner, and Dov Gabbay. "Extracting and Tagging Unstructured Citation of a Hebrew Religious Document." Pages 461–73 in \textit{Proceedings of the Information Science and Information Technology Education Conference}, 2019.

Naaijer, Martijn, and Dirk Roorda. "Parallel Texts in the Hebrew Bible, New Methods and Visualizations." (2016). https://hal.science/hal-01283051v1.

Naaijer, Martijn, Constantijn Sikkel, Mathias Coeckelbergs, Jisk Attema, and Willem Van Peursen. "A Transformer-Based Parser for Syriac Morphology." Pages 23–29 in \textit{Proceedings of the Ancient Language Processing Workshop Associated with the RANLP-2023}. Association for Computational Linguistics, 2023.

Okuda, Nozomu. "Deep Learning for Intertext Discovery in Latin Epic: Latin SBERT and the Lucan-Vergil Benchmark." ProQuest Dissertations and Theses. Ph.D., State University of New York at Buffalo, 2022.

Ramdas, Aaditya, Nicolás Trillos, and Marco Cuturi. "On Wasserstein Two-Sample Testing and Related Families of Nonparametric Tests." \textit{Entropy} 19/2 (2017): 47.

Reimers, Nils, and Iryna Gurevych. "Reporting Score Distributions Makes a Difference: Performance Study of LSTM-Networks for Sequence Tagging." Pages 338–48 in \textit{Proceedings of the 2017 Conference on Empirical Methods in Natural Language Processing}. Copenhagen: Association for Computational Linguistics, 2017.

\authdash. "Sentence-BERT: Sentence Embeddings Using Siamese BERT-Networks." Pages 3980–90 in \textit{Proceedings of the 2019 Conference on Empirical Methods in Natural Language Processing and the 9th International Joint Conference on Natural Language Processing (EMNLP-IJCNLP)}. Association for Computational Linguistics, 2019.

Reshef, Yael. \textit{Historical Continuity in the Emergence of Modern Hebrew}. Lanham, MD: Lexington Books, 2019.

Rodrigue Schwarzwald, Ora. "Lexicon: Modern Hebrew." \textit{Encyclopedia of Hebrew Language and Linguistics Online}: Brill, 2013.

\authdash. "Morphology: Modern Hebrew." \textit{Encyclopedia of Hebrew Language and Linguistics Online}: Brill, 2013.

\authdash. "The Components of the Hebrew Lexicon: The Influence of Hebrew Classical Sources, Jewish Languages and Other Foreign Languages on Modern Hebrew." \textit{Hebrew Linguistics} 39 (1995): 79–90.

Roorda, Dirk, Gino Kalkman, Martijn Naaijer, and Andreas van Cranenburgh. "LAF-Fabric: A Data Analysis Tool for Linguistic Annotation Framework with an Application to the Hebrew Bible." \textit{arXiv:1410.0286}, 1 October 2014. \url{https://doi.org/10.48550/arXiv.1410.0286}

Rosensweig, Elisha, Benjamin Resnick, Hillel Gershuni, Joshua Guedalia, Nachum Dershowitz, and Avi Shmidman. "Automatic Text Segmentation of Ancient and Historic Hebrew." Pages 1–11 in \textit{Proceedings of the Second Workshop on Ancient Language Processing}. Association for Computational Linguistics, 2025.

Schmid, Friedrich, and Axel Schmidt. "Nonparametric Estimation of the Coefficient of Overlapping—Theory and Empirical Application." \textit{Computational Statistics \& Data Analysis }50 (2006): 1583–1596.

Seker, Amit, Elron Bandel, Dan Bareket, Idan Brusilovsky, Refael Shaked Greenfeld, and Reut Tsarfaty. "AlephBERT: A Hebrew Large Pre-Trained Language Model to Start-off Your Hebrew NLP Application With." \textit{arXiv:2104.04052}, 8 April 2021. \url{https://doi.org/10.48550/arXiv.2104.04052}

Shmidman, Avi, and Moshe Koppel. "Torah Study and the Digital Revolution: A Glimpse of the Future." \textit{The }\textit{Lehrhaus}, 28 January 2020. \url{https://thelehrhaus.com/commentary/torah-study-and-the-digital-revolution-a-glimpse-of-the-future/}.

Shmidman, Avi, Joshua Guedalia, Shaltiel Shmidman, Cheyn Shmuel Shmidman, Eli Handel, and Moshe Koppel. "Introducing BEREL: BERT Embeddings for Rabbinic-Encoded Language."\textit{ arXiv:2208.01875}, 2022. \url{https://doi.org/10.48550/arXiv.2208.01875}

Shmidman, Avi, Moshe Koppel, and Ely Porat. "Identification of Parallel Passages Across a Large Hebrew/Aramaic Corpus."\textit{ Journal of Data Mining \& Digital Humanities Special Issue on Computer-Aided Processing of Intertextuality in Ancient Languages} (2018): 1388.

Shmidman, Avi, Ometz Shmidman, Hillel Gershuni, and Moshe Koppel. "MsBERT: A New Model for the Reconstruction of Lacunae in Hebrew Manuscripts." Pages 13–18 in \textit{Proceedings of the 1st Workshop on Machine Learning for Ancient Languages (ML4AL 2024)}. Association for Computational Linguistics, 2024.

Shmidman, Avi, Shaltiel Shmidman, Moshe Koppel, and Yoav Goldberg. "Nakdan: Professional Hebrew Diacritizer." Pages 197–203 in \textit{Proceedings of the 58th Annual Meeting of the Association for Computational Linguistics: System Demonstrations.} Association for Computational Linguistics, 2020.

Shmidman, Avi. "Automatic Identification of Biblical Citations and Allusions in Hebrew Texts." Pages 335–48 in \textit{Jewish Studies in the Digital Age}. De Gruyter, 2022.

Shmidman, Shaltiel, Avi Shmidman, and Moshe Koppel. "DictaBERT: A State-of-the-Art BERT Suite for Modern Hebrew." \textit{arXiv:2308.16687}, 13 October 2023. \url{https://doi.org/10.48550/arXiv.2308.16687}

Shmidman, Shaltiel, Avi Shmidman, Amir DN Cohen, and Moshe Koppel. "Adapting LLMs to Hebrew: Unveiling DictaLM 2.0 with Enhanced Vocabulary and Instruction Capabilities." \textit{arXiv:2407.07080}, 9 July 2024. \url{https://doi.org/10.48550/arXiv.2407.07080}

Shmidman, Shaltiel, Avi Shmidman, Moshe Koppel, and Reut Tsarfaty. "MRL Parsing Without Tears: The Case of Hebrew." Pages 4537–50 in \textit{Findings of the Association for Computational Linguistics ACL 2024}. Association for Computational Linguistics, 2024.

Sivan, Reuven. \textit{The Revival of the Hebrew Language}. Jerusalem: Rubinstein, 1980.

Smiley, David M. "Intertextual Parallel Detection in Biblical Hebrew: A Transformer-Based Benchmark." \textit{arXiv:2506.24117}, 30 June 2025. \url{https://doi.org/10.48550/arXiv.2506.24117}

\authdash. ``T'OMIM: Tanakh Observable Matches of Intertextual Mimesis''. \textit{Zenodo}, March 22, 2026. \url{https://doi.org/10.5281/zenodo.19135731}

Tan, Qitao, Xiaoying Song, Guanghui Ye, and Chuan Wu. "An Effective Negative Sampling Approach for Contrastive Learning of Sentence Embedding." \textit{Machine Learning} 112 (2023): 4837–61.

Thakur, Nandan, Nils Reimers, Andreas Rücklé, Abhishek Srivastava, and Iryna Gurevych. "BEIR: A Heterogeneous Benchmark for Zero-shot Evaluation of Information Retrieval Models." Pages 1–16 in \textit{Proceedings of the 2021 Conference on Neural Information Processing Systems: Datasets and Benchmarks Track}. NeurIPS, 2021.

Tschuggnall, Michael, and Günther Specht. "From Plagiarism Detection to Bible Analysis: The Potential of Machine Learning for Grammar-Based Text Analysis." Pages 245–48 in \textit{Joint European Conference on Machine Learning and Knowledge Discovery in Databases}. Cham: Springer International Publishing, 2016.

Tsumura, David Toshio. \textit{Vertical Grammar of Parallelism in Biblical Hebrew.} Ancient Israel and Its Literature 47. Atlanta: SBL Press, 2023.

van Peursen, Willem, and Eep Talstra. "Computer-Assisted Analysis of Parallel Texts in the Bible. The Case of 2 Kings Xviii-Xix and Its Parallels in Isaiah and Chronicles." \textit{Vetus Testamentum} 57/1 (2007): 45–72.

van Peursen, Willem, Constantijn Sikkel, and Dirk Roorda. "Hebrew Text Database ETCBC4b." Eep Talstra Centre for Bible and Computing, VU University Amsterdam, and Data Archiving and Networked Services, Royal Netherlands Academy of Arts and Sciences, 2015.

Vaswani, Ashish, Noam Shazeer, Niki Parmar, Jakob Uszkoreit, Llion Jones, Aidan N. Gomez, Lukasz Kaiser, and Illia Polosukhin. "Attention Is All You Need." Pages 6000–6010 in \textit{Proceedings of the 31st Conference on Neural Information Processing Systems (NIPS)}. Long Beach, CA: Curran Associates, 2017.

Wang, Hao, and Yong Dou. "SNCSE: Contrastive Learning for Unsupervised Sentence Embedding with Soft Negative Samples." Pages 419–31 in \textit{Advanced Intelligent Computing Technology and Applications.} Singapore: Springer Nature Singapore, 2023.

Wang, Jiapeng, and Yihong Dong. "Measurement of Text Similarity: A Survey." \textit{Information} 11/9 (2020): 421.

Woods, Peter J. "Shitposting as Public Pedagogy." \textit{Curriculum Inquiry} 53/4 (2023): 359–80.

Xu, Lanling, Jianxun Lian, Wayne Xin Zhao, Ming Gong, Linjun Shou, Daxin Jiang, Xing Xie, and Ji-Rong Wen. "Negative Sampling for Contrastive Representation Learning: A Review."\textit{ arXiv:2206.00212}, 2022. \url{https://doi.org/10.48550/arXiv.2206.00212}

Xu, Lingling, Haoran Xie, Zongxi Li, Fu Lee Wang, Weiming Wang, and Qing Li. "Contrastive Learning Models for Sentence Representations." \textit{ACM Transactions on Intelligent Systems and Technology} 14/4 (2023): 1–34.

Yamini, Bat-Zion. "The Revival of Ancient Hebrew Words With the Revival of Israel." \textit{Sociology Study} 9/4 (2019): 156–168.

Yang, Zhen, Ming Ding, Tinglin Huang, Yukuo Cen, Junshuai Song, Bin Xu, Yuxiao Dong, and Jie Tang. "Does Negative Sampling Matter? A Review with Insights into Its Theory and Applications."\textit{ IEEE Transactions on Pattern Analysis and Machine Intelligence} 46/8 (2024): 5692–711.

\end{hangparas}

\end{document}